\documentclass[10pt,journal,compsoc]{IEEEtran}

 \usepackage{cite}
\usepackage{graphicx}
\usepackage{subcaption}
\usepackage{booktabs} 
\usepackage{hyperref}
\usepackage{bbding}
\usepackage{bm}
\usepackage{lipsum}

\usepackage{enumitem}

\usepackage{amsmath}
\usepackage{amssymb}
\usepackage{mathtools}
\usepackage{amsthm}
\usepackage{comment}
\usepackage[capitalize,noabbrev]{cleveref}

\usepackage{algorithmic}
\usepackage{times}
\usepackage{epsfig}
\usepackage{graphicx}
\usepackage{amsmath}
\usepackage{amssymb}
\usepackage{booktabs}
\usepackage{diagbox}
\usepackage{booktabs}
\usepackage[linesnumbered,ruled,vlined]{algorithm2e}
\SetKwInput{KwData}{Input}
\SetKwInput{KwResult}{Output}
\usepackage{comment}
\usepackage{times}
\usepackage{epsfig}
\usepackage{graphicx}
\usepackage{amsmath}
\usepackage{arydshln}
\usepackage{amssymb}
\usepackage{multirow}
\usepackage{color}
\usepackage{framed}
\theoremstyle{plain}
\newtheorem{theorem}{Theorem}[section]

\theoremstyle{definition}
\newtheorem{definition}[theorem]{Definition}

\theoremstyle{remark}

\usepackage[textsize=tiny]{todonotes}
\usepackage{caption}
\begin{document}

\title{HDGT: Heterogeneous Driving Graph Transformer for Multi-Agent Trajectory Prediction via Scene Encoding}

\author{Xiaosong Jia,~\IEEEmembership{Student~Member,~IEEE},
        Penghao Wu,~
        Li Chen,~
        Yu Liu,~
        Hongyang Li,~\IEEEmembership{Senior~Member,~IEEE},
        ~Junchi~Yan,~\IEEEmembership{Senior~Member,~IEEE}
 \IEEEcompsocitemizethanks{\IEEEcompsocthanksitem X. Jia, P. Wu, H. Li and J. Yan are with Department of Computer Science and Engineering, and MoE Key Lab of Artificial Intelligence, Shanghai Jiao Tong University, and Shanghai AI Lab, Shanghai, China.\protect\\
 L. Chen and Y. Liu are with Shanghai AI Lab, Shanghai, China.\protect\\
This work was partly done when X. Jia and P. Wu were interns at the Shanghai AI Lab.\protect\\

E-mail: \{jiaxiaosong,wupenghaocraig,yanjunchi\}@sjtu.edu.cn\protect\\
{\{lichen, liuyu, lihongyang\}@pjlab.org.cn},
\protect\\
Correspondence author: Junchi Yan\protect
}}
\markboth{}%
{Shell \MakeLowercase{\textit{et al.}}: HDGT: Heterogeneous Driving Graph Transformer for Multi-Agent Trajectory Prediction via Scene Encoding}
\IEEEtitleabstractindextext{
\begin{abstract}
Encoding a driving scene into vector representations has been an essential task for autonomous driving that can benefit downstream tasks e.g. trajectory prediction. The driving scene often involves heterogeneous elements such as the different types of objects (agents, lanes, traffic signs) and the semantic relations between objects are rich and diverse. Meanwhile, there also exist relativity across elements, which means that the spatial relation is a relative concept and need be encoded in a ego-centric manner instead of in a global coordinate system. Based on these observations, we propose Heterogeneous Driving Graph Transformer (HDGT), a backbone modelling the driving scene as a heterogeneous graph with different types of nodes and edges. For heterogeneous graph construction, we connect different types of nodes according to diverse semantic relations. For spatial relation encoding, the coordinates of the node as well as its in-edges are in the local node-centric coordinate system. For the aggregation module in the graph neural network (GNN), we adopt the transformer structure in a hierarchical way to fit the heterogeneous nature of inputs. Experimental results show that HDGT achieves state-of-the-art performance for the task of trajectory prediction, on INTERACTION Prediction Challenge and Waymo Open Motion Challenge.
\end{abstract}}

\maketitle

\section{Introduction}\label{sec:intro}
Autonomous driving is one of the most influential domains for AI applications in recent years. 
An autonomous driving system is made of a few key modules:
perception/localization, prediction, planning,  and control. The perception/localization module is responsible to extract visual and semantic information from raw sensor data and HD-Map. 
The outcomes of preceding functionalities are in compact yet highly unstructured forms, e.g. positions and states of surrounding vehicles, pedestrians and traffic lights, and relations between lanes. 
For downstream tasks such as trajectory prediction and planning, it is essential to obtain comprehensive semantic representations from those unstructured inputs. 
The conventional pipeline for the task is to first acquire a representation vector for each target agent, which should exhibit both ego moving patterns and states of surrounding environment. 
Then given the vector as input, the subsequent decoder could conduct prediction/planning in the desired form (raw trajectory/control signal/etc), combined with some prior assumptions or physical constraints. 
However, even deeply inspired by the success of deep learning literature in recent years, the encoding process needs to be investigated nonetheless.
There is no de facto paradigm to encode a representative vector due to the following observations.
\begin{figure}[!tb]
\centering
\includegraphics[width=8cm]{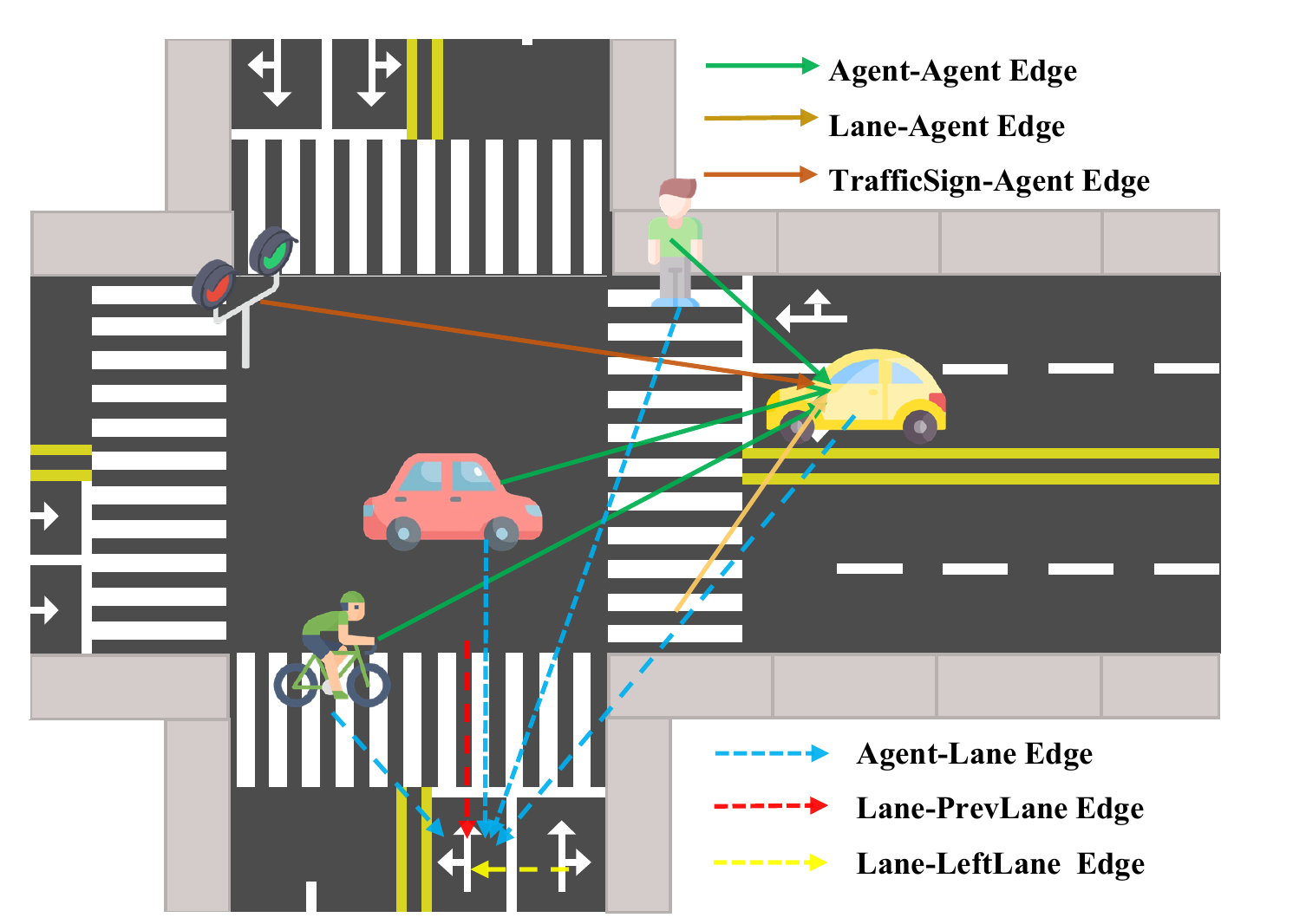}
\caption{Illustration of a heterogeneous driving graph. In the proposed framework, different types of agents and map elements are modeled as different types of nodes. The diverse semantic relations among them are modeled as different types of edges.}
\label{fig:teaser}
\end{figure}

One inherent property that lies in the data is the \textit{heterogeneity}, where we imply objects and their relations to have different types. The input data may include agents' motion states, traffic lights with position, lane polylines with feasible movement connections, road line polygons with various types, \textit{etc}.
Fig.~\ref{fig:teaser} illustrates an heterogeneous driving  graph.
It is non-trivial to encode those information comprehensively since they are highly unstructured, compared to the image input in the computer vision (CV) or the text input in the natural language processing (NLP).
Thus, instead of directly the adopting widely used CNN or Transformers, \emph{a good network design should consider a wide span of the input types and their semantic relations separately and explicitly}.

There has been recent progress towards exploiting the heterogeneity of the driving scene.
Trajectron++~\cite{DBLP:journals/corr/abs-2001-03093} and HEAT-R~\cite{mo2021heterogeneous} encode different types of agents with different parameters 
but they encode the map elements by rasterized representation, which leads to inferior performance compared to vector-based methods~\cite{Gao_2020_CVPR}.
VectorNet based approaches~\cite{Gao_2020_CVPR,zhao2020tnt,DBLP:journals/corr/abs-2108-09640,DBLP:journals/corr/abs-2109-06446} and SceneTransformer~\cite{ngiam2021scene} put all agents and lanes in one single fully-connected graph and use shared parameters for information aggregation, which ignores different node types and semantic relations. 
LaneGCN based solutions~\cite{LiangECCV20,zeng2021lanercnn,deo2021multimodal,gilles2021thomas,DBLP:journals/corr/abs-2109-01827} introduce a series of four graphs (LaneToAgent, LaneToLane, LaneToActor, and finally ActorToActor) according to distance heuristics and connective information of the lane. 
It demonstrates better performance compared to VectorNet and its extensions~\cite{Gao_2020_CVPR,zhao2020tnt}, which suggests the importance of modeling heterogeneity. 
However, the sequential order of those four different relation aggregation functions in LaneGCN is designed manually by intuition.

The aforementioned issues make LaneGCN-like models not very flexible in face of datasets with different size, which makes it hard to simply apply the modern stacking-scaling~\cite{devlin2018bert,NEURIPS2020_1457c0d6,dosovitskiy2020vit} principle to increase the capacity of the model with more layers in the industry. To this end, we propose the \textbf{H}eterogeneous \textbf{D}riving \textbf{G}raph \textbf{T}ransformer, namely \textbf{HDGT}, where the driving scene is modeled as a heterogeneous graph and with different types of nodes (agents or map elements) and the edges encode the the semantic relations between two nodes.

Specifically, we devise the aggregation and update function of the nodes and edges in a spirit from the recent success of Transformer~\cite{vaswani2017attention}. To capture the heterogeneity, different sets of Transformer parameters are adopted for different node and edge types.
Different from LaneGCN-like models, all nodes and edges are updated simultaneously. The same structure (Transformer) and simultaneous updating of HDGT make it easy to stack and scale when dealing with different data types and sizes.
We conduct thorough ablation studies to demonstrate the importance of exploiting heterogeneity. We conduct experiments to verify the scalability of HDGT under different datasets with different size of models.

Besides heterogeneity, another import property of the driving scene data is the \textit{relativity}, where we indicate each agent in the driving scene should process surrounding elements in its local coordinate system.
If we adopt global reference, models would simply overfit the value of the absolute coordinates on the training scenes, and perform poorly on the unseen scenes. 
For example, to predict a vehicle's future trajectory, the model should consider whether it would go straight or turn left or turn right, which is in relative reference instead of a global one.
Hence, instead of processing the absolute coordinates, it is more reasonable to unanimously encode the relative relations among elements in a local view. Such a view shifts from other elements to the agent need be incorporated into the network explicitly.
Most existing works choose a reference point~\cite{zhao2020tnt,LiangECCV20,ngiam2021scene} (for example, the ego agent) and then all the other elements including vehicles and lanes are in the reference point's coordinate system.
Though this way achieves relativity for the ego agent, we note that the representations of non-ego elements are not in their local coordinate systems. Consequently, as suggested in~\cite{ettinger2021large}, to obtain the best results for multi-agent prediction, those methods have to forward multiple times with each agent as the reference agent, which brings notable additional cost.
In this paper, \emph{we aim to design a model where all agents` are processing information in their local coordinate system and they are treated symmetrically so that predictions for the entire scene could be done in a single forward.}

To this end, in our proposed HDGT approach, the spatial features of nodes as well as their in-edges\footnote{In this paper, the heterogeneous graph we build is directional, which means the edge feature from
$A$ to $B$ is different from $B$ to $A$. We denote in-edges of $A$ as all the edges going into $A$.} are in each node's local, ego-centric system. As a result, the aggregation and updating of node features are done in their own reference, which captures the relative spatial state of other objects when their orientation is upfront and position is (0, 0). Then, for an edge $(u \rightarrow v)$, since its feature represents u's states in $v$'s coordinate system, we update it with the node feature of $u$ (which is in u's coordinate system) combining with the coordinate transformation information from $u$ to $v$. In the aforementioned two steps, we update node features in their local coordinate system and edge features back to its in-node's coordinate system, which exploits the spatial relativity between elements and keeps all nodes and edges symmetric.
Ablation studies in the experiment part will demonstrate the effectiveness of the proposed design.

Towards a practical multi-agent trajectory prediction engine for autonomous driving, the main highlights of this paper are:
\begin{itemize}
    \item We devise an unified heterogeneous GNN for trajectory forecasting with HD-map and propose a hierarchical way of adopting Transformer to fit the heterogeneous nature of the inputs. The explicit modeling of all semantics and relations in the scene could improve the prediction accuracy by a large margin. The unified Transformer structure and simultaneous updating of all elements make it easy to stack and scale.
    \item Instead of choosing one reference agent, in a similar spirit with  concurrent works~\cite{varadarajan2021multipath++,zhou2022hivt}, we propose a symmetric way of encoding the spatial relationship between elements in the scene, which notably improves the generalization ability and enables the model to predict multiple agents' future in one forward without performance drop.
    \item We examine the proposed HDGT on two challenging large-scale datasets and it achieves state-of-the-art performance on the corresponding online leaderboards. Thorough ablation studies are performed to evaluate the effectiveness of each designed component.
    
\end{itemize}

In particular, our extensive experimental results show that the proposed HDGT achieves new records on two challenging benchmarks, namely INTERACTION Prediction Challenge~\cite{zhan2019interaction} and Waymo Open Motion Challenge~\cite{ettinger2021large}. Throughout 11/3/2021 (our final submission to INTERACTION) to 4/21/2022 (we published this report in arxiv), we still ranked the $\mathbf{1^{st}}$ and $\mathbf{2^{nd}}$ respectively in terms of minADE/minFDE metrics. Source code and trained models are publicly available at: \url{https://github.com/OpenPerceptionX/HDGT}.

\section{Related Work}
\label{sec:related}
\subsection{Heterogeneous Graph Neural Networks}
\subsubsection{General Heterogeneous GNNs}
Graph, as a common structure, is composed of nodes and edges, where a node usually represents an entity or element and a edge represents some kind of relation between the two connected nodes. To encode the structure and information of a graph, Graph neural network (GNN)~\cite{DBLP:conf/iclr/KipfW17} is proposed, which usually contains an aggregation function and an update function. The aggregation function is applied on each node/edge to collect information from their neighbors while the update function is to generate new representations for each node/edge according to their collected information and old representations. Two most widely used models are GCN~\cite{DBLP:conf/iclr/KipfW17} and GAT~\cite{velivckovic2018graph}, which adopts the Laplacian matrix of the graph and attention for aggregation and updating respectively. For graphs with different types of nodes and edges, heterogeneous graph neural network is further introduced. R-GCN~\cite{10.1007/978-3-319-93417-4_38} uses GCN~\cite{DBLP:conf/iclr/KipfW17} like aggregating functions but with different parameters for different types, while HGN~\cite{wang2019heterogeneous} and HetSANN~\cite{Hong_Guo_Lin_Yang_Li_Ye_2020} adopt GAT~\cite{velivckovic2018graph} like aggregation function with different parameters for different types. HetGNN~\cite{zhang2019heterogeneous} adopts the random walk to generate neighbors for nodes and uses distinct RNNs for different types of nodes. HGT~\cite{hu2020heterogeneous} apply different Transformers for different relations on the academic graph.

\subsubsection{Heterogeneous GNNs for Trajectory Prediction\label{sec:hgnn-for-traj}}
In the heterogeneous graph of a driving scene, a node could either be an agent (vehicle, pedestrian, cyclist, etc) or a map element (lane, traffic light, stop sign, etc). A node usually contains some information (called node feature) such as agents' state (position, velocity), lane's type and the corresponding composing polyline, etc. One edge usually contains information (called edge feature) e.g. relative position between two entities or relations (Lane-LeftLane, Agent-Lane, etc). Trajectron++~\cite{DBLP:journals/corr/abs-2001-03093} and HEAT-R~\cite{mo2021heterogeneous} encode different types of agents with different parameters of GNNs or Transformer. However, as pioneering works, they encode the map elements by rasterized images and use a CNN to extract representation vectors of maps. According to the recent progress in the trajectory prediction community, it is inefficient and leads to much inferior performance compared to vector-based methods~\cite{Gao_2020_CVPR}. In fact, vector-based methods have dominated leaderboards of multiple public datasets Waymo~\cite{ettinger2021large}, nuScenes~\cite{nuscenes}, Argoverse~\cite{chang2019argoverse} and INTERACTION~\cite{zhan2019interaction}. Different from existing works, in HDGT, we constructs one heterogeneous graph including both agents and map elements and an unified Transformer structure with type-specific parameters is adopted for the aggregation and update function of heterogeneous graph.

\subsection{Transformer}
\subsubsection{General Transformers}
Transformer~\cite{vaswani2017attention}, as a set function to aggregate information of elements, has show promising success in multiple fields. The two core components of Transformer is the dot-product attention and the feed-forward network. The dot-product attention mechanism computes the pair-wise relativity of two elements by dot-product and updates the representations of each element by the normalized weighted sum of other elements.  Emerging from the NLP field, BERT~\cite{devlin2018bert} and GPT series~\cite{radford2019language,brown2020language} demonstrate the strong scalability of Transformers when handling huge amount of data, which could benefit from large scale pretraining. Later, Transformer is also shown to be able to scale well on image data, which starts the era of Vision Transformer (ViT)~\cite{DBLP:journals/corr/abs-2010-11929}. Besides, in fields including speech~\cite{baevski2020wav2vec}, multimodal learning~\cite{kim2021vilt}, reinforcement learning~\cite{chen2021decision}, and etc, Transformer also achieves SOTA performance, which exhibits its wide applicability.

\subsubsection{Transformers for Trajectory Prediction}
Transformers have also recently been adopted in autonomous driving, especially for trajectory prediction. However, their map elements are separately encoded thorough rasterized representation by a CNN, which loses parts of heterogeneity. VectorNet encoder~\cite{Gao_2020_CVPR} adopts attention mechanism on a fully connected graph with agents and map elements and SceneTransformer~\cite{ngiam2021scene} additionally apply attention on the temporal axis. IDE-Net~\cite{DBLP:journals/corr/abs-2011-02403} explicitly extracts the interaction between agents by Transformer. However, the above works basically update representations of all agents and map elements with the same set of sharing parameters, ignoring the heterogeneity of nodes and relations. In our HDGT, instead of building a fully connected graph and sharing parameters for all nodes, we connect nodes by their semantic relation and conduct aggregation and updating of node and edge embeddings with type-specific parameters. Additionally, all nodes are processing information in their local coordinate systems. Experiments show that HDGT outperforms VectorNet/DenseTNT~\cite{DBLP:journals/corr/abs-2108-09640} and SceneTransformer~\cite{ngiam2021scene} by a notable margin.

\begin{figure*}[t!]
\centering
\includegraphics[width=.96\textwidth]{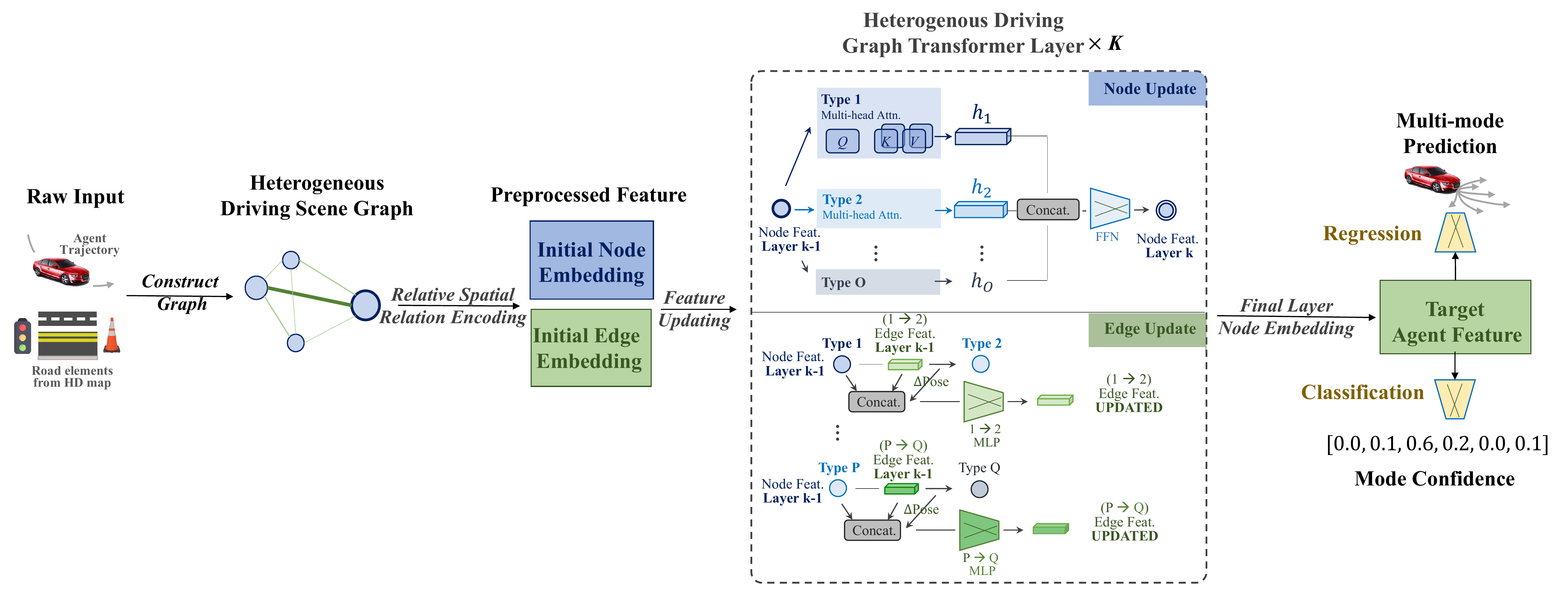}
\caption{\textbf{The pipeline of HDGT.} We first construct heterogeneous driving scene graph according to agents' and map elements' types and relations. Then, we initialize the node and edge feature of the heterogeneous graph by relative spatial encoding. Next, we update the node and edge features by the Heterogeneous Driving Graph Transformer Layer. The updating is conducted $K$ times by $K$ layers. Finally, we take the target agent nodes' feature at the last layer and feed them the regression and classification MLP to obtain the multi-modal trajectory prediction with confidence scroes.
}\label{fig:Pipeline}
\end{figure*}

\subsection{Trajectory Prediction for Autonomous Driving}
\subsubsection{Scene Encoder For Trajectory Prediction}
Pioneering methods for trajectory prediction focus on the interaction among agents (called Social Force~\cite{helbing1995social}) and do not take the HD-Map into consideration. Social LSTM~\cite{Alahi_2016_CVPR} combines LSTM with social pooling. Social GAN~\cite{Gupta2018SocialGS} computes relative positions between the ego and all other people, rather than only considering people inside the grid as did in Social pooling. The authors in~\cite{ivanovic2018generative} consider the edge relations to better capture the interaction. The work~\cite{ivanovic2018generative} designs a GNN for agents detected from raw sensor data to capture their interactions. Trajectron~\cite{ivanovic2019trajectron} proposes a graph-structured encoder for the ease of predicting distributions of multi-agent trajectory. ST-GCN~\cite{mohamed2020social} adopts GCN on both spatial and temporal dimension. 

However, in the driving scene, the movement of agents is not solely influenced by other agents. The map elements including lanes, traffic lights, and stop signs could all have significant influence on agents' future trajectories. Thus, it is necessary to take the map into consideration in a prediction model. Since the HD-map data is highly unstructured (polylines and their types), for deep learning based methods, one popular choice is to use the rasterized images to represent the scene~\cite{tang2019multiple,DBLP:journals/corr/LeeCVCTC17,zhao2019multi,chai2019multipath,cui2019multimodal,hong2019rules}. Specifically, they project map elements into a top-down view image according to the 2D coordinates; different types of elements may be painted in different channels. In this way, the convolutional neural network (CNN) could directly applied on the rasterized image to extract the map information. As for the encoder for agents, one choice is to apply RNN and GNN similar to pioneering works and then combine it with the rasterized representation of map. Following this branch,  Trajectron++~\cite{DBLP:journals/corr/abs-2001-03093} improves Trajectron with the rasterized HD-Map and heterogeneous agents type. \cite{mo2021heterogeneous} builds an agent-only heterogeneous graph neural network whose edges including the relative information. \cite{marchetti2020mantra} designs a memory network to store the scene knowledge. \cite{choi2021shared} uses multiple sensors' data for trajectory prediction. Different from 
the aforementioned works, ~\cite{bansal2018chauffeurnet} adopts the rasterization for agents as well and further formulates the output on the rasterized images.  With the blossom of CNN structure in vision, this raster-based  methods could outperform those without map information. 

Nevertheless, as demonstrated in \cite{Gao_2020_CVPR}, the raster-based methods lose too much information during painting and are inefficient to generate good representation, since the road elements are sparse in image space. Besides, it is difficult for CNN to capture global information due to its limited effective perception field issue~\cite{luo2016understanding}. Most recently, encoding the map in a vector-based spirit \cite{Gao_2020_CVPR,LiangECCV20} has proven great performance in multiple competitions including Waymo~\cite{ettinger2021large}, nuScenes~\cite{nuscenes}, Argoverse~\cite{chang2019argoverse} and INTERACTION~\cite{zhan2019interaction}. Works in this branch typically employ 1D CNN or LSTM \cite{HochSchm97} to process the temporal data, adopt the PointNet~\cite{Qi_2017_CVPR} to process polylines, and use graph neural network (GNN) to handle the relation among all agents and elements.   VectorNet~\cite{Gao_2020_CVPR} uses sub-graph for lane and agent encoding and a fully connected global graph to capture their relations. SceneTransformer~\cite{ngiam2021scene} proposes a factorized spatial-temporal network which applies Transformer on the fully-connected spatial/temporal graph alternatively.  LaneGCN~\cite{LiangECCV20} introduces a series of four different information aggregations modules for the driving scene. TPCN~\cite{ye2021tpcn} treats all coordinate data as point clouds and achieves comparable performance with SOTA works. 
\subsubsection{Output Decoder for Trajectory Prediction}

Besides those works focusing on extracting high-quality representations of the scene (i.e. the encoder part), there are also works explore the output formulation and training paradigm for the trajectory prediction, i.e., the decoder part.  Social LSTM~\cite{Alahi_2016_CVPR} assumes an independent bivariate Gaussian distribution for the future position of each time-step of each agent. Social GAN~\cite{Gupta2018SocialGS} further introduces GAN to improve the diversity of prediction. Trajectron~\cite{ivanovic2019trajectron} combines recurrent sequence modeling and variational deep generative modeling. Trajectron++~\cite{DBLP:journals/corr/abs-2001-03093} additionally discusses the paradigm of future-conditional predictions. Prank~\cite{DBLP:journals/corr/abs-2010-12007} produces the conditional distribution of agent’s trajectories plausible in the given scene.~\cite{casas2020implicit} aims to model  the joint distribution over future trajectories via an implicit latent variable model with GNN for agents only. TNT~\cite{zhao2020tnt} introduces goal-based predictions for diversity and DenseTNT~\cite{DBLP:journals/corr/abs-2108-09640} further designs dense goal sets and a corresponding optimization pipeline to further improve its performance. LaneRCNN~\cite{zeng2021lanercnn} proposes local LaneROI and based on it goal-based prediction is used.  ~\cite{deo2021multimodal} proposes to predict multi-modal future conditioned on lane connecting topology. ~\cite{DBLP:journals/corr/abs-2109-01827} proposes the graph-based heatmap output paradigm to optimize the missing rate while ~\cite{gilles2021thomas} designs a hierarchical heatmap paradigm and a learnable trajectory recombination method for joint multi-agent trajectory prediction. DCMS~\cite{ye2022dcms} introduces dual consistency and multi-pseudo-target supervision for training. ~\cite{pmlr-v205-jia23a} conducts refinements with temporal priors.

In this work, we focus on the encoder part, which aims to extract the massive diverse information from the scene comprehensively. Comparing to the existing works, we are the first to comprehensively model all node types and all semantic and geometric relations into one single heterogeneous graph and update the node and edge features in a unanimous and parallel way but with distinct parameters for different node or edge types. Additionally, we propose to process all nodes in their own local coordinate system, which enhances the generalization ability of the model and enables the single-forward multi-agent prediction feasible. Notably, concurrent works Multipath++~\cite{varadarajan2021multipath++}\footnote{Specifically, Multipath++~\cite{varadarajan2021multipath++}'s preprint  was initially uploaded to Arxiv on 29/11/2021. While we submitted our results to INTERACTION Leaderboard on 03/11/2021 as recorded on the INTERACTION Leaderboard.} has similar designs where all the features in each agent's local coordinate system. HiVT~\cite{zhou2022hivt} proposes to process the scene in a hierarchical way with a local part and a global interaction module by Transformer. With similar ideas of the local coordinate system, they both achieve new records on the public leaderboards, which demonstrates importance of relativity as mentioned in Sec.~\ref{sec:intro}.

\section{Methodology}
\label{sec:method}

\subsection{Problem Formulation and Approach Overview}
We introduce our method. The input of the trajectory prediction includes two parts: agents' history states and map elements. Agents' history states is composed of past $L$ time-steps in a certain frequency (for example, 10 Hz) where at each time-step agents' coordinate, velocity, heading. etc are provided. Besides, the size of bounding box (width, length, height) and the type (pedestrian, cyclist, vehicle) of agents are given as well. Map elements are surrounding objects and signs which reflect traffic rules and thus have nonnegligible influence on agents' future movements. The commonly given map elements include: lane, road line, crosswalk, stop sign, traffic light, speed bump, curb, etc. Each map element contains a polyline or a polygon to indicate its physical position as well as other information: for traffic lights, their state (green, yellow, red) at each time-step are given; for traffic lights and stop signs, the lanes which they have influence on are marked; for lanes, their predecessors, successors, and neighbors are given.

The goal of the trajectory prediction task is to predict the future trajectories of agents in the next $T$ time-steps. To better capture the uncertainty of predicting future, multiple different predictions together with confidence scores are allowed for each agent.

The pipeline of HDGT is shown in Fig.~\ref{fig:Pipeline}, which involves: 

\begin{itemize}
    \item Sec. \ref{sec:construction} describes how we construct the heterogeneous graph of the driving scene based on the raw input.
    \item Sec. \ref{sec:relative} gives how we transform the the node and edge feature into local reference system.
    \item Sec. \ref{sec:graph} describes how Transformer structure is adopted to aggregate and update the node and edge features.
    \item Sec. \ref{sec:loss} describes the heads and loss used to do the prediction based on the extracted agent node features. 
\end{itemize}

\subsection{Construction of the Heterogeneous Graph}\label{sec:construction}
\textbf{Preliminary}
In this work, we represent each driving scene with multiple agents and map elements as one directed heterogeneous graph.
A directed graph is represented as $\mathcal{G} = \{\mathcal{V}, \mathcal{E}\}$, where $\mathcal{V}$ denotes the set of nodes and $\mathcal{E}$ denotes the set of edges. 
A heterogeneous graph has a node type
mapping function of $\tau(v): \mathcal{V} \rightarrow \text{Type}_v$ and an edge type mapping function $\phi(e): \mathcal{E} \rightarrow \text{Type}_e$, where $\text{Type}_v$ and $\text{Type}_e$ denote the set of node and edge type respectively. 
In HDGT, the type of an edge $e: u \rightarrow v$ is determined by the type of source node $u$ and destination node $v$ and their relations.
When building the heterogeneous graph, all possible combinations of semantics relations are 
considered.

For an agent node, instead of connecting it to all the other counterparts like VectorNet~\cite{Gao_2020_CVPR}, LaneGCN\cite{LiangECCV20} constrains each node only connecting to those whose distance is smaller than a threshold. 
However, in some scenarios, vehicles traversed a long way in the prediction horizon. For example, in Waymo Open Motion, some vehicles move more than 300 meters in the 8 seconds prediction horizon.
It is computationally expensive and unnecessary to set a fixed large threshold for \textit{all} agents. For example, for pedestrians, vehicles very far away would not influence their future behaviors.

For all agents' node, we calculate their distance with others by their position at the final observed time-step; for all nodes with the polyline shape, we calculate by the middle point of the polyline; for all nodes with a polygon shape, we calculate by their geometric centre.
Thus, similar to~\cite{zeng2021lanercnn}, for each agent, we set the distance threshold flexible to be its speed times the prediction horizon with a type-specific buffer value. 

For a lane or traffic signs node, we inversely connect them to the agent nodes according to the threshold mentioned in the paragraph above. Besides, for lane nodes, we connect them with their surrounding lane nodes and there are four types - left, right, entry, and exit. Additionally, since the length of each lane could vary a lot, we divide those lanes whose length is larger than a threshold into short ones with similar length so that all lane nodes have similar inputs. 
We give the pseudo code for the construction of the driving scene heterogeneous graph in Alg.~\ref{alg:construct}

Accordingly, the node types include agents, lanes, and traffic signs (stop signs, road lines, cross walks). The edge types include Agent-Lane, Agent-Trafficsign, Lane-Agent, Lane-NextLane, Lane-PreviousLane, Lane-LeftLane, Lane-RightLane, and Trafficsign-Agent.

\begin{algorithm}[tb!]
\caption{Construct Heterogeneous Graph for Driving Scene Encoding\label{alg:construct}}
\DontPrintSemicolon

\KwData{$N_a$ agents` position, $N_l$ lanes` positions and their connective topology, $N_{tr}$ traffic signs positions}

\KwResult{
Node set $\mathcal{V}$, adjacency matrix $\mathbf{A}\in \mathbb{R}^{|\mathcal{V}|\times |\mathcal{V}|}$
}

\Begin{
Divide lanes with length larger than threshold $\epsilon_\text{lane}$ into [$\frac{\text{length}}{ \epsilon_\text{lane}}$] pieces.

Node set $\mathcal{V} \gets$ Agents $\cup$  Lanes  $\cup$ Traffic Signs

Initialize Adjacency Matrix $\mathbf{A}\in \mathbb{R}^{|\mathcal{V}|\times |\mathcal{V}|}$ with zeros.

\For{$i=1,2,\cdots, N_a$}{
    Retrieve agent node $v_{a, i}$'s position $\text{pos}_{a, i}$, distance threshold $\epsilon_{a, i}$, and index $\text{idx}_{a, i}$ in $\mathcal{V}$
    
    \For{$j=1,2,\cdots, |\mathcal{V}|$}{
         Retrieve node $v_{j}$'s position $\text{pos}_{j}$
        
        \If{distance($\text{pos}_{a,i}$, $\text{pos}_{j}$) $< \epsilon_{a,i} $}{
            $\mathbf{A}[\text{idx}_{a, i}, j] \gets$   {Relation}(Agent-Type(j));
            
            $\mathbf{A}[j, \text{idx}_{a, i}] \gets$   {Relation}(Type(j)-Agent);
        }
}}

\For{$i=1,2,\cdots, N_l$}{
    Retrieve lane node $v_{l, i}$'s connective topology $Connected$ and index $\text{idx}_{l, i}$ in $\mathcal{V}$;
    
    \For{Lane node $v_{l, j}$ $\in$ Connected($v_{l, i}$)}{
        Retrieve $v_{l, j}$'s connective relation with $v_{l, i}$ - $CR$ and index $\text{idx}_{l, i}$ in $\mathcal{V}$;
        
        $\mathbf{A}[\text{idx}_{l, i}, \text{idx}_{l, j}] \gets $  Relation(Lane-$CR$);
    }
}

return $\mathcal{V}$, $\mathbf{A}$;
}
\end{algorithm}

\begin{figure}[tb!]
\centering
\includegraphics[width=.45\textwidth]{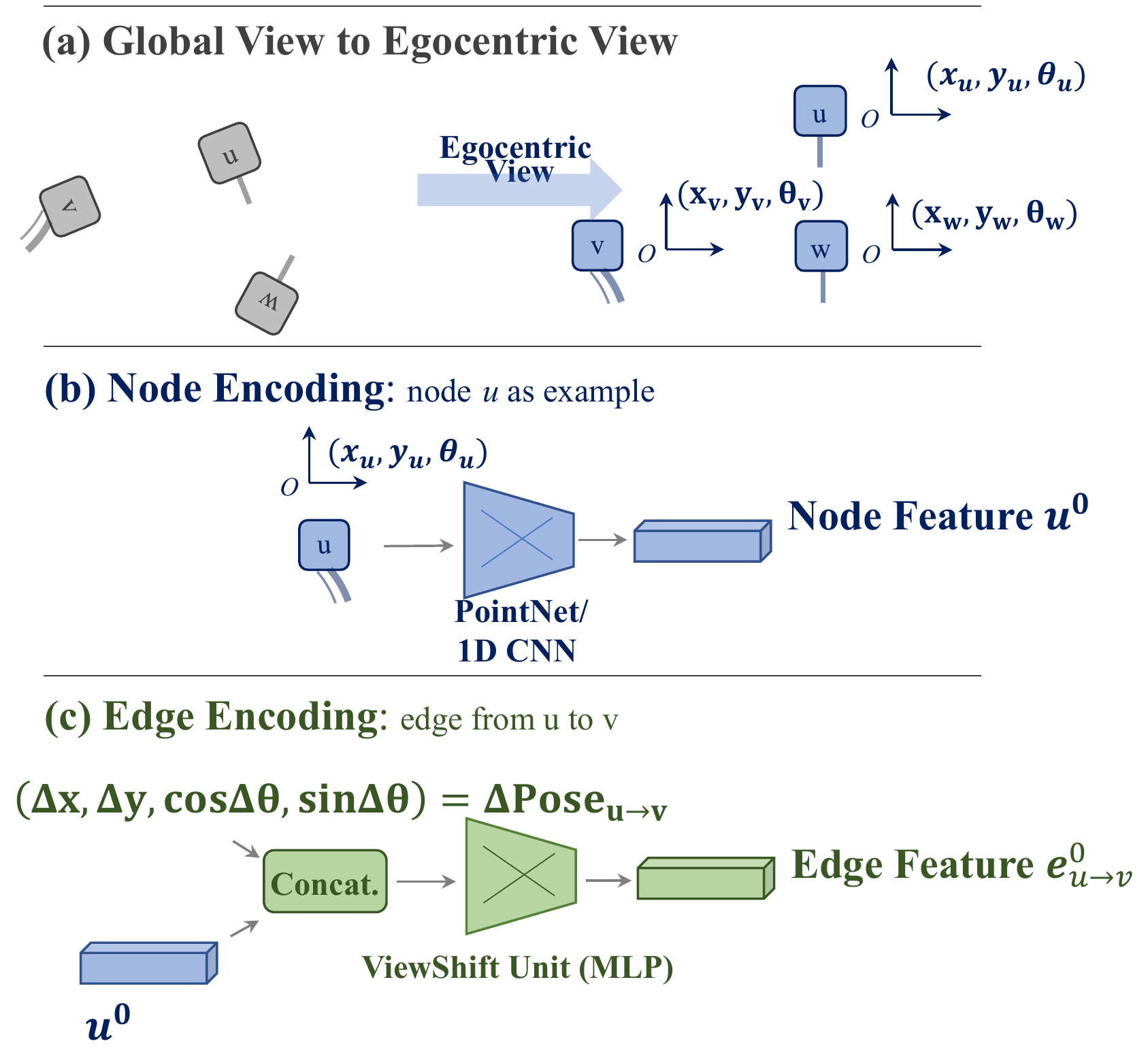}
\caption{Process of the Relative Spatial Relation Encoding which encodes both node and edge features in their own local reference system. (a) Spatial information encoding of the node, where we transform all nodes in their ego-centric coordinate system. (b) To transform node feature into a single compact vector instead of one trajectory/polyline/polygon, we adopt 1D CNN or PointNet. (c) Transform source node‘s representation vector into target node's in-edge feature by concatenating with the delta between two coordinate systems and feeding them into an MLP.}\label{fig:relative}
\end{figure}

\subsection{Relative Spatial Relation Encoding}\label{sec:relative}

After constructing the heterogeneous graph of the driving scene, we need to initialize the feature embedding of all nodes and edges.

As mentioned in Sec.~\ref{sec:intro}, it is undesirable to directly feed raw data to the neural network since the raw spatial data are usually in a global coordinate system defined by datasets. Considering same set of spatial relations could be represented differently by the coordinate system with different origins and orientations, the spatial data should be normalized into a canonical form.
Otherwise, non-normalized inputs would inappropriately force the network to overfit the absolute value of training data and thus unfavorably cause poor generalization.
To normalize the data, one popular choice is to select one agent as a global reference~\cite{Gao_2020_CVPR,LiangECCV20}. 
Though this way achieves relativity for the ego agent, we note that the representations of non-ego elements are not in their local coordinate systems. 
For example, assume there are three vehicles in the scene. If we choose vehicle 1's current coordinate as the origin and heading as the positive y-axis, then in vehicle 1's view, all vehicles moves relative to its current state. However, in vehicle 2 and vehicle 3's view, they observe other vehicle's moving in a biased coordinate system which is un-normalized and could lead to worse generalization ability.
Consequently, as suggested in~\cite{ettinger2021large}, to obtain the best results for multi-agent prediction, those methods have to forward multiple times with each agent as the reference agent, which brings notable additional cost.
Thus, we should design a model where all agents are processing information in their local coordinate system so that they could all be treated treated symmetrically into a canonical form. In this way, predictions for the entire scene could be done in a single forward.

For nodes which represent either agents or map elements, we initialize their feature embeddings in their own ego-centric coordinate system. Specifically, for each node $v$, we define its unique reference as a tuple $(x^v_\text{ref}, y^v_\text{ref}, \theta^v_\text{ref})$ where $x^v_\text{ref}$ and $y^v_\text{ref}$ represents the origin of the coordinate system and $\theta^v_\text{ref}$ represents the orientation. For all coordinates feature of node v, we subtract $\begin{bmatrix} x \\ y \end{bmatrix}$ by $\begin{bmatrix} x^v_\text{ref} \\ y^v_\text{ref} \end{bmatrix}$ so that all these coordinates are relative to itself. Then, for all coordinate and velocity features, we rotate them by $-\theta^v_\text{ref}$, i.e., we let the rotation matrix $R=\begin{bmatrix}
\cos-\theta^v_\text{ref} & -\sin-\theta^v_\text{ref}\\
\sin-\theta^v_\text{ref} & \cos-\theta^v_\text{ref}
\end{bmatrix} = \begin{bmatrix}
\cos\theta^v_\text{ref} & \sin\theta^v_\text{ref}\\
-\sin\theta^v_\text{ref} & \cos\theta^v_\text{ref}
\end{bmatrix}$ dot product with either $\begin{bmatrix} x \\ y \end{bmatrix}$ or $\begin{bmatrix} \text{velocity}_x \\ \text{velocity}_y \end{bmatrix}$. Besides, for agents' node with heading feature, we subtract the them with $\theta^v_\text{ref}$. In this way, all the node features are composed of information relative to itself. Note that for agent nodes, we set $(x^v_\text{ref}, y^v_\text{ref})$ as its position at the final observed time-step and $\theta^v_\text{ref})$ as its heading at the final observed time-step, which means each agents' history trajectory and velocity are relative to its current state. During training, for labels - future trajectories of agent, we conduct the transformation as well so that the regression target are relative as well. For map elements composed of a polyline, we select the middle point's position as $(x^v_\text{ref}, y^v_\text{ref})$ and the direction of the vector from its start point to its end point as the $\theta^v_\text{ref}$. For map elements composed of a polygon, we set $(x^v_\text{ref}, y^v_\text{ref})$ as the  geometric centre of the polygon and  $\theta^v_\text{ref}$ as the direction of the longest edge of the polygon. In this way, all the map elements are represented in a canonical form. Fig.~\ref{fig:relative}(a) illustrates how we transform all node features in their local reference system.

Until now, all node features are in their ego-centric coordinate system. However, they are still unstructured, i.e., they could be a sequence of history states for agents, a polyline or a polygon. For the ease of efficient parallel operations in the graph neural network, we transform them into single vectors with the same length first. Following the common practice in~\cite{Gao_2020_CVPR,LiangECCV20}: for all agent nodes, we adopt a shared 1D-CNN on each of their features to encode their temporal dynamics. The 1D CNN is composed of a series of 1D convolution layers, downsampling and finally pooling to obtain a single vector representation. For all map element nodes, we adopt a simplified version of PointNet. Specifically, for a set of points from either a polyline or a polygon, a  PointNet layer is composed of a shared MLP for each point, an max-pooling operation over all points to obtain the overall representation, and finally concatenating the pooled vector with the vector after MLP. By stacking several layers of PointNet (here we set 3 layers) and finally adopting a max-pooling operation, the geometry information of a polyline or a polygon is encoded in a single vector. Fig.~\ref{fig:relative}(b) shows the process that we encode node features in their ego-centric view by 1D CNN or PointNet. Additionally, sub-type information such as type of lanes (single, double, etc) and road lines (broke-broke, solid-broke, etc), we adopt learnable embeddings for them and then concatenate them with the geometry feature obtained from 1D CNN or PointNet. We denote the initial node feature set as $\{\mathbf{v}^0\}$ for all nodes $v$ in $\mathcal{V}$.

For edges in the directed graph, we initialize their feature embeddings in their target nodes` coordinate system. Formally, for an edge $u \rightarrow v$, we refer $u$ as its origin node and $v$ as its target node. The edge feature represents node $u$'s information in the view of node $v$. For example, one edge feature $e_{u \rightarrow v}$ might represent  a lane $u$ in the left in agent $v$'s view. Recall that we already have the representation vector of $u$, as shown in Fig.~\ref{fig:relative}(b). However, the vector of node $u$ - $u^0$, is in $u$'s local coordinate system instead of $v$'s. Thus, we need to transform $u^0$ into the local coordinate system of $v$. Denote the reference of node $u$ and $v$ as $(x^u_\text{ref}, y^u_\text{ref}, \theta^u_\text{ref})$ and $(x^v_\text{ref}, y^v_\text{ref}, \theta^v_\text{ref})$ respectively. To transform coordinate from $u$ to $v$'s coordinate system, the required information is $(x^v_\text{ref}-x^u_\text{ref}, y^v_\text{ref}-y^u_\text{ref}, \cos(\theta^v_\text{ref}-\theta^u_\text{ref}), \sin(\theta^v_\text{ref}-\theta^u_\text{ref})) = (\Delta x_{u\rightarrow v}, \Delta y_{u\rightarrow v}, \cos\Delta\theta_{u\rightarrow v}, \sin\Delta\theta_{u\rightarrow v})$, denoted as $\Delta \text{Pose}_{u\rightarrow v}$. For simpleness, we adopt MLP layers to obtain the initial edge feature $e_{u\rightarrow v}^0$ by:
\begin{equation}
    \mathbf{e}_{u\rightarrow v}^0 = 
     {\operatorname{MLP}}_{\tau(u)} \left(\texttt{concat} \left[\mathbf{u}^0, \Delta \text{Pose}_{u\rightarrow v} \right] \right)
 \label{equ:e-mlp}
\end{equation}
where $r(u)$ means that the MLP used for the transformation depends on the type of source node $u$ considering the fact that different node types (agent, lane, traffic sign) have different geometry characteristics in their node features.
In this way, we provide in-edge features of $v$ with information related to $v$'s coordinate system and this operation is symmetric for all nodes.
We denote this MLP as \emph{ViewShift Unit} as demonstrated in Fig.~\ref{fig:relative}(c)

In a nutshell, by the relative spatial relation encoding procedure mentioned in this section, we decouple the spatial encoding process into two parts: First, the motion/geometric characteristics of a single object -- node feature encoding Second, the spatial relation between two objects -- edge feature encoding. As a result, the inputs for the neural network are all in the ego-centric coordinate system, which satisfies the i.i.d input assumption and is less prone to overfitting absolute coordinate value. To be more specific, assume two vehicles in the same scene have exactly the same patterns. One vehicle moves from (0, 0) to (5, 5) and the other moves from (100, 100) to (105, 105). Choosing any vehicle as the reference would cause the encoder having to deal with inputs at different scales while the encoder still needs to output similar features since they share the same patterns. In the proposed way, the node features are exactly the same while the in-edge features contain the relative relation. We could observe the symmetry within both node features and edge features, which is beneficial for the model's generalization ability since the same type of inputs share parameters.

\subsection{Heterogeneous Driving Graph Transformer Layer}\label{sec:graph}
After building the heterogeneous driving graph and initialize node and edge features with relative spatial encoding, the next step is to conduct information exchange among nodes and edges and update their corresponding features. In this way, the complex relations among agents and map elements could be encoded so that we could feed the predicted agents' node feature to the decoder head to generate their future multi-mode trajectories. 

\subsubsection{Preliminaries\label{sec:gnn-transformer-preliminary}}
\textbf{Transformer}: Transformer~\cite{vaswani2017attention} is a permutation-invariant set function which can fuse the information among a set in pair-wise way by dot-product attention. It mainly consists of two parts: the multi-head attention module to fuse information from others and the feed-forward module to update information of themselves. Formally, suppose we have a set of $m$  features with dimension $h$: $\mathbf{Z}=\{z_i|i=1,2,...,m\} \in \mathbb{R}^{m\times h}$.
In the attention mechanism, three linear transformations ($\mathbf{W}_q \in \mathbb{R}^{h\times h_q}, \mathbf{W}_k \in \mathbb{R}^{h\times h_k}, \mathbf{W}_v \in \mathbb{R}^{h\times h_v}, h_q=h_k$) are conducted to generate the three vectors for each element in the set, called query (Q), Key (K), Value (V):
\begin{equation}
    \mathbf{Q}, \mathbf{K}, \mathbf{V} = \mathbf{Z}\mathbf{W}_q, \mathbf{Z}\mathbf{W}_k, \mathbf{Z}\mathbf{W}_v
\end{equation}
Then, a dot-product operation followed by a Softmax operation is conducted between $\mathbf{Q}$ and $\mathbf{K}$ to generate pair-wise importance weights $\mathbf{A} \in \mathbb{R}^{m\times m}$ which is called attention matrix:
\begin{equation}
    \mathbf{A} = \operatorname{Softmax}(\mathbf{Q}\mathbf{K}^\top/\sqrt{h_q})
\end{equation}
where $\sqrt{h_q}$ is for the numerical stability of the Softmax.
Finally, the weighted sum is performed with weights $\mathbf{A}$ and value $\mathbf{V}$:
\begin{equation}
    \mathbf{Z}^\prime = \mathbf{A}\mathbf{V}
\end{equation}
Thus, we define the attention function as $\mathbf{Z}^\prime = \text{Attn}(\mathbf{z})$. The multi-head attention (MHA) extends the attention by applying multiple attention on the same set of inputs under different parameters (called different heads) in parallel and then applying a linear transformation, namely $\mathbf{W}_o \in \mathbb{R}^{n_hh_v\times h_v}$, where $n_h$ is the number of heads, to integrate the heads:
\begin{equation}
    \mathbf{Z}^{\prime \prime} = \operatorname{MHA}(\mathbf{Z}) = [\operatorname{Attn}_1(\mathbf{\mathbf{Z}}),...,\operatorname{Attn}_{n_h}(\mathbf{\mathbf{Z}})]W_o
\end{equation}
The feed-forward module is a 2-layer MLP with activation $\delta(\cdot)$:
\begin{equation}
    \mathbf{Z}^{\prime \prime \prime} = \operatorname{FFN}(\mathbf{Z}^{\prime \prime}) = \delta(\mathbf{Z} \mathbf{W}_1+\mathbf{b}_1)\mathbf{W}_2+\mathbf{b}_2
\end{equation}
The $\mathbf{Z}^{\prime \prime \prime}$ is the output of one Transformer layer and it usually shares exactly the same hidden dimension with $\mathbf{Z}$ for the ease of adding residual connection. Note that within MHA and FFA, residual connections~\cite{he2016deep} and layer norm (LN)~\cite{ba2016layer} are applied for the stability of training:
\begin{equation}
    \begin{split}
        & \mathbf{Z}^{\prime\prime} =  \operatorname{MHA}(\operatorname{LN}(\mathbf{Z})) + \mathbf{Z} \\
        & \mathbf{Z}^{\prime\prime\prime} =  \operatorname{FFN}(\operatorname{LN}(\mathbf{Z}^{\prime\prime})) + \mathbf{Z}^{\prime\prime}
    \end{split}
\end{equation}

\hspace{-5mm}\textbf{Graph Neural Network (GNN)}: in HDGT, the information exchange and feature updating are fulfilled by GNN. We give the definition of the aggregation function and update function~\cite{battaglia2018relational}.
\begin{definition}
\textbf{Aggregation Function}: An aggregation function $\rho$ is a map which takes a set as input, and reduces it to a single element which represents the aggregated information. The aggregation functions must be invariant to permutations of their inputs, and should take variable numbers of arguments (e.g., element-wise summation, mean, maximum, etc).
\end{definition}

\begin{definition}
\textbf{Update Function}: An update function $\phi$ is a map conducted on all elements in the same set (for example, all agent nodes is a set while all lane nodes is another set) and serves as an integration of information from ego and neighbors.
\end{definition}

In HDGT, for the heterogeneous graph, there are two types of aggregation function and two types of update function: $\rho^{\mathcal{V}}$ is applied on nodes and aggregates features from each node's in-edges; $\rho^{\mathcal{E}}$ is applied on edges and aggregates features from each edge's source node. $\phi^{\mathcal{V}}$ and $\phi^{\mathcal{E}}$ update node and edge features respectively based on aggregated information and their current features.

In the following sections, we introduce how we adopt Transformer as the aggregation and update function of GNN in the spirit of heterogeneity and relativity. In HDGT, there are $K$ layers of GNN, which means the node and edge updating mentioned below are conducted $K$ times to extract the high-order relation among agents and map elements.

\subsubsection{Node Aggregation and Update}

For the aggregation function of nodes  $\rho^{\mathcal{V}}$, we adopt the multi-head attention (MHA) of Transformer to collect information from each node's in-edges which is their own ego-centric coordinate system. However, Transformer is initially designed for language data, which assumes all elements are homogeneous - words. 
This assumption does not hold true for the driving scene data since in-edges could be agents (dynamic states) or lanes (connective topology) or traffic signs (traffic rules).
Therefore, it is unreasonable to let all in-edges share the same set of parameters for the aggregation function.
Additionally, since an agent node's neighbor is decided by its speed as well as a buffer value, we find that some agent nodes may have thousands of in-edges. Consequently, due to the sparse nature of the attention mechanism, it is difficult to aggregate information from rare but significant nodes (for example, crosswalk) and thus leads to degenerated performance. 

To alleviate the aforementioned issues, we propose a hierarchical manner to adopt Transformer by first processing different types of semantic relations separately by attention and then integrating information from different types with an MLP.
For the aggregation of in-edges' features, each type of in-edges shares a set of parameters of MHA and does cross-attention by letting the node features as $\mathbf{Q}$ and the in-edge features as $\mathbf{K}$ and $\mathbf{V}$ 
As a result, there would be a set of attended features - each for a specific edge type. 
Finally, we feed the concatenation of the set into an MLP to obtain the aggregated features of each node.

Formally, for a node $v$ with a set of $m$ in-edges $\{e_i\}_{i=0}^{i=m}$ in the $k^{th}$ layer, the input is node feature $\mathbf{v}^{k-1}$ and its in-edge features $\{\mathbf{e}^{k-1}_i\}_{i=0}^{i=m}$ from the last layer. Note that for layer 1, we use the features initialized by relative spatial relation encoding in Sec.~\ref{sec:relative}. Denote the features of all in-edges of $v$ with a specific edge type $j$ from $(k-1)^{th}$ layer as $\mathbf{E}^{k-1}_j \in \mathbb{R}^{m_j \times h}$ where $m_j$ is the number of type-$j$ in-edges and $h$ is the hidden dimension, we calculate the attention vector of this specific type for this node as:
\begin{equation}
    \mathbf{a}_j = \operatorname{Softmax} (\mathbf{q}_j\mathbf{K}_j)^\top/\sqrt{h} ), \in \mathbb{R}^{m_j}
    \label{equ:attn}
\end{equation}
where $\mathbf{q}_j = {\mathbf{v}}^{k-1}\mathbf{W}_{j, \text{query}} \in \mathbb{R}^{h}$, $\mathbf{K}_j = \mathbf{E}_j\mathbf{W}_{j, \text{key}} \in \mathbb{R}^{m_j \times h}$, $\mathbf{W}_{j, \text{query}}$ and $\mathbf{W}_{j, \text{key}}$ are the query and key linear transformation matrix for edge type $j$. In this way, the attention score is calculated in a type-specific way, which utilizes the prior of heterogeneity. For example, for an agent to drive safely, its attention for other agents and for lanes should be different and treated separately.

For all types of relations, we conduct MHA with different parameters in the aforementioned way, which outputs a set of aggregated features $\{\mathbf{v}^{k-1,\prime\prime}_j \in \mathbb{R}^{h} | j=1,...,n_r\}$ where $n_r$ is the number of possible in-edge types for $v$'s node type. To integrate them, an MLP layer is adopted to obtain the final aggregated node feature $\mathbf{v}^{k-1,\prime\prime}$ as:
\begin{equation}
    \mathbf{v}^{k-1,\prime\prime}  = {\rm{MLP}}\left(\mathop{\rm \texttt{concat}}\left(\{\mathbf{v}^{k-1,\prime\prime}_j | j=1,...,n_r\}\right)
    \right)
    \label{equ:cat-mlp}
\end{equation}
Here, the operation of the aggregation function $\rho^{\mathcal{V}}$ is finished.

As for the update function of nodes $\phi^{\mathcal{V}}$, we adopt FFN to update all node features. Considering the heterogeneous nature of nodes, we adopt different parameters for different types of nodes:
\begin{equation}
    \mathbf{v}^{k+1} = \text{FFN}_{\text{Type}(v)}(\mathbf{v}^{k-1,\prime\prime})
\end{equation}
where $\mathbf{v}^{k+1}$ is the updated features of node $v$. Note that for the clarity  of the expression, we do no write the residual connection and layer norm layers in the equations above while we do adopt them following~\cite{vaswani2017attention}.

\subsubsection{Edge Aggregation and Update}

Recall that edge feature $\mathbf{e}_{u \rightarrow v}$ represents element $u$'s representation in element $v$'s local view. Thus, to update $\mathbf{e}_{u \rightarrow v}$, we need to combine node $u$'s feature and transform it to node $v$'s view.

Formally, assume at layer $k$ we have $\mathbf{u}^{k-1}$, $\mathbf{v}^{k-1}$, and $\mathbf{e}^{k-1}_{u \rightarrow v}$ which represents the node feature of u and v and their edge feature on last layer respectively. Note that for layer 1, we use the features initialized by relative spatial relation encoding in Sec.~\ref{sec:relative}. 

For the aggregation function of edge $\rho^{\mathcal{E}}$, we retrieve node $u's$ feature $\mathbf{u}^{k-1}$ and the transform between node $u$ and $v$'s coordinate system $\Delta \text{Pose}_{u\rightarrow v}$ as defined in Sec.~\ref{sec:relative}.

For the update function of edge $\phi^{\mathcal{E}}$, we concatenate the aggregated information $\mathbf{u}^{k-1}$ and $\Delta \text{Pose}_{u\rightarrow v}$  with its previous feature $\mathbf{e}^{k-1}_{u \rightarrow v}$ and feed them into an MLP to obtain the updated edge feature $\mathbf{e}^{k}_{u \rightarrow v}$. Again, for different edge types, we adopt different MLPs hence the heterogeneity is ensured:
\begin{equation}
    \mathbf{e}^{k}_{u \rightarrow v} = {\rm MLP}_{\text{type}(u \rightarrow v)} \left( \operatorname{concat} \left[\mathbf{u}^{k-1}, \Delta \text{Pose}_{u\rightarrow v}, \mathbf{e}^{k-1}_{u \rightarrow v}\right] \right)
\end{equation}
Similarly, we adopt the residual connection and layer norm in the update function as well and we do not write it for brevity.

In summary, after conduct the aforementioned operations for all nodes and edges for once, higher order of relation information with semantics could be extracted. Note that all aggregation functions and update functions are similar to the ones used in Transformer~\cite{vaswani2017attention}, which enables the model to exploit the scalability by simply stacking layers while the modification for heterogeneity and relativity injects the task-specific prior into the structure. In Fig.~\ref{fig:Pipeline}, Heterogeneous Driving Graph Transformer Layer part gives an visualization about how we conduct node and edge feature updating at each layer.

\subsection{Output Head and Training Objective}\label{sec:loss}
For the output head, we adopt the commonly used regression-MLP plus classification-MLP combination. It falls in the formulation of the MTP loss~\cite{cui2019multimodal}. Different from the original one used in~\cite{cui2019multimodal} which selects the winner trajectory by a distance function that considers an angle between the last
points of the two trajectories as seen from the actor position, we follow the recent practice in~\cite{Gao_2020_CVPR,LiangECCV20} and adopt the average Euclid distance as the selection criterion for the winner.
Specifically, for each target agent, we use the node feature at HDGT's last layer as its hidden representation. An MLP module is used for regression; it outputs a tensor of size $N \times K \times T \times 2$, where $N$ is the number of target agents, $K$ is the number of required mode, $T$ is the prediction length, and 2 corresponds to the dimension size of $(x, y)$. 
Another MLP is used to generate the confidence of each mode for each agent - a tensor of size  $N \times K$. 
For different types of agents (vehicles/pedestrians/cyclists), different parameters for the two heads are used to capture the diverse moving behaviors.
Denote an agent's outputs as:
\begin{align}
    O_{\rm{reg}} & = \{(\mathbf{p}^k_{1}, \mathbf{p}^k_{2}, \cdots \mathbf{p}^k_{T})\}_{k \in [1, K]},
    \\
    O_{\rm{cls}} & = \{c^k\}_{k \in [1, K]},
\end{align}
where $\mathbf{p}^k_{i} = (x^k_i, y^k_i)$ is the predicted coordinate of this agent at time step $i$ in the $k^{th}$ mode, 
and $c_k$ is the confidence.

As for the training objective, for each agent, we only backpropagate loss for the regression head with the lowest regression loss to avoid mode collapse and encourage diversity. For the classification head, we use the cross-entropy loss. The mode with the lowest regression loss is set as the positive sample, and the rest modes are set as negative samples consequently.

Mathematically, for the $n^{th}$ agent, we find the index $k_n$ of predicted mode with lowest regression loss:
\begin{equation}
    k_{n} = \mathop{\rm argmin}_{k \in [1, K]}\frac{1}{2T} \sum_{t=1}^T \left[d(x^k_{n, t}, x^*_{n, t}) + d(y^k_{n, t}, y^*_{n, t})\right],
\end{equation}
where $x^*_{n, t}$, $y^*_{n, t}$ is the ground truth coordinate at time step $t$ of the $n^{th}$ agent, and we adopt the widely used smooth L1 loss as~\cite{girshick2015fast}:
\begin{equation}
    d(x_1, x_2) =
\begin{cases}
0.5(x_1-x_2)^2& \text{if}\, \|x_1-x_2\|_1 < 1, \\
\|x_1-x_2\|_1 - 0.5& \text{otherwise}.
\end{cases}
\end{equation}
The final training objective can be calculated as:
\begin{equation}
\begin{aligned}
    \mathcal{L} &= \lambda\mathcal{L}_{\rm cls} + \mathcal{L}_{\rm reg} \\
                &= \lambda\frac1N \sum_{n=1}^N \operatorname{ CrossEntropy}\left(\{c^k\}_n, I(k_{n})\right) \\
                &~~~~+ \frac{1}{2NT}\sum_{n=1}^N\sum_{t=1}^T \left[d(x^{k_n}_{n, t}, x^*_{n, t}) + d(y^{k_n}_{n, t}, y^*_{n, t})\right],
\end{aligned}
\end{equation}
where $I(k_{n})$ is an one-hot vector of length $K$ whose $k_n^{th}$ entry is 1 and $\lambda$ is the hyperparameter set to be 0.1 throughout the paper.

\section{Experiments}
We depict the experimental results in details. Sec.~\ref{sec:dataset} describes the datasets, metrics and the implementation details; Compared with the peer methods in Sec. \ref{sec:compare_baselines}, the numerical analysis is provided in Sec. \ref{sec:sota} where state-of-the-art works are compared. Ablation studies are given in Sec.~\ref{sec:ablation} with more detailed study and comparison of different Relative Encoding Strategies om Sec.~\ref{sec:comparision-relative} and  Heterogeneous Graph Design in Sec.~\ref{sec:het-graph-design}. Sec.~\ref{sec:scale} investigates the model capacity of HDGT. The qualitative visualization is presented subsequently in Sec.~\ref{sec:visualize} to give the audience a glimpse of what the model learns in various complicated scenarios.

\subsection{Protocols}\label{sec:dataset}
\subsubsection{Metrics}
Similar to~\cite{Gao_2020_CVPR,LiangECCV20}, we use minADE, minFDE, and MR as the validation metric. Note that different benchmarks have slightly different definitions of the aforementioned metrics. Some other metrics such as  minJointADE (minSADE), minJointFDE  (minSFDE), SMR~\cite{zhan2019interaction,casas2020implicit}, and mAP~\cite{ettinger2021large} are also used. Therefore, we use the following metrics in the validation set and report the leaderboard results in spirit of the official metrics respectively.

Formally, denote the multi-mode prediction of the model as $O = \{(\mathbf{p}^k_{1}, \mathbf{p}^k_{2}, \cdots \mathbf{p}^k_{T})\}_{k \in [1, K]}$
where $\mathbf{p}^k_{t} = (x^k_t, y^k_t)$ is the predicted coordinate of the agent at time step $t$ in the $k^{th}$ mode. Denote the ground-truth trajectory of the agent as $O^* = (\mathbf{p}^*_{1}, \mathbf{p}^*_{2}, \cdots \mathbf{p}^*_{T})$.

\begin{enumerate}
    \item \textbf{minADE} (Minimum Average Displacement Error): the minimum value of the Euclidean distance between the prediction and ground truth averaged by the prediction length $T$, for $K$ required predictions. It measures how good the predictions are in an average spirit under the Euclidean space. The detailed calculation is:
    \begin{equation}
        \text{minADE} = \min_{k=1,\cdots,K} \frac1T\sum\limits_{t=1}^T ||\mathbf{p}^k_t-\mathbf{p}^*_t||_2
    \end{equation}
    
    \item \textbf{minFDE} (Minimum Final Displacement Error): similar to minADE, despite that  it only calculates the error at the final time-step $T$. It focuses more on the accuracy of the predicted goal point, which emphasizes on the importance of predicting agents' intentions. The detailed calculation is:
    \begin{equation}
        \text{minFDE} = \min_{k=1,\cdots,K}  ||\mathbf{p}^k_T-\mathbf{p}^*_T||_2
    \end{equation}
    
    \item \textbf{MR} (Missing Rate): the ratio of whether the Euclidean distance between the prediction and ground truth at the final time-step $T$ for all $K$ predictions is larger than 2m. Different from Euclid-based metrics like minADE and minFDE, it only requires the model to predict the coarse-grained intention. If any one of the predicted mode falls in the neighborhood of the ground-truth final point, it is counted as a hit (a 0 for the miss rate). The detailed calculation is:
    \begin{equation}
        \text{MR} = \left\{
        \begin{array}{ll}
        0, \quad\quad  \exists k\in\{1,...,K\}, ||\mathbf{p}^k_T-\mathbf{p}^*_T||_2 \leq 2 &\\
        1, \quad\quad \text{Otherwise}
        \end{array}\right.
    \end{equation}
\end{enumerate}

Note that all three aforementioned metrics is averaged over all target agents.

\subsubsection{Datasets}
We use two popular and large-scale datasets for evaluation.
\begin{enumerate}
\item \textbf{INTERACTION Dataset} It consists of various highly interactive driving situations, including highway ramps, roundabouts, and intersections, recorded using drones or fixed cameras worldwide~\cite{zhan2019interaction}. We use its most recent 1.2 version which provides the official train/val/test split and the labels of test set are held out for the online competition. In this dataset, target agents are only vehicles, which has 40K tracks in total. The input data is 1 second history in 10 Hz and the required output is 3 seconds future in 10Hz. At most 6 modes are allowed.
Note that there are four tracks in the INTERACTION challenges with different data and settings. We adopt their settings by: for the two conditional tracks which additionally give the future trajectory of the ego agent, we ignore the additional information for the consistency with other datasets and experiments. For the two joint tracks which require joint multi-agent prediction, we let all predictions with the same index as one modality.

\item \textbf{Waymo Open Motion Dataset} 
It provides a large scale dataset with annotations for objects with interacting behaviors over a wide range of road geometries~\cite{ettinger2021large} over a diverse set of locations, with 1750km unique roadways. It contains 7.64M agent tracks in total. We use its most recent 1.2 version with road connection information. It also provides the official train/val/test split and has an online leaderboard for test set. The target agents are divided into 3 types: vehicles, pedestrians, cyclists. The input is 1.1 seconds history in 10 Hz and required output is 8 seconds future in 2Hz. At most 6 modes are allowed. They provide up to 8 objects to predict in each scene, where their selection of target is biased to require objects that do not follow a constant velocity model or straight paths. We follow the official train/val/test split of the datasets above and report the test set results on the online leaderboards. 
\end{enumerate}

\begin{table*}[tb!]
\footnotesize
\caption{\textbf{Results on the INTERACTION Leaderboard, \textit{test} set.} }
\begin{subtable}{0.5\textwidth}
\centering
\begin{tabular}{llll}
 Method                     & minADE$\downarrow$ & minFDE$\downarrow$ & MR$\downarrow$     \\ \hline
DenseTNT~\cite{DBLP:journals/corr/abs-2108-09640}               & 0.4342 & 0.7952 & 0.0596 \\
MultiModalTransformer~\cite{DBLP:journals/corr/abs-2109-06446} & 0.2130 & 0.5511 & 0.0511 \\
GOHOME~\cite{DBLP:journals/corr/abs-2109-01827}                & 0.2005 & 0.5988 & \textbf{0.0491} \\
\hline
\textbf{HDGT (Ours)}            & \textbf{0.1676} & \textbf{0.4776} & 0.0556
\end{tabular}
\caption{Single Agent Track}
\end{subtable}
\begin{subtable}{0.5\textwidth}
\centering
\begin{tabular}{llll}
 Method                     & minADE$\downarrow$ & minFDE$\downarrow$ & MR$\downarrow$     \\ \hline
DenseTNT~\cite{DBLP:journals/corr/abs-2108-09640}               & 0.5452 & 0.6375 & 0.0348 \\
GOHOME~\cite{DBLP:journals/corr/abs-2109-01827} & 0.2020 & 0.5902 & 0.0284 \\
MultiModalTransformer~\cite{DBLP:journals/corr/abs-2109-06446} & 0.1989 & 0.5799 & 0.0429 \\
\hline
\textbf{HDGT (Ours)}            & \textbf{0.1553} & \textbf{0.4577} & \textbf{0.0276}
\end{tabular}
\caption{Conditional Single Agent Track}
\end{subtable}
\begin{subtable}{0.5\textwidth}
\centering
\begin{tabular}{llll}
 Method                     & minSADE$\downarrow$ & minSFDE$\downarrow$ & SMR$\downarrow$     \\ \hline
ReCoG2~\cite{mo2020recog}             & 0.4668 & 1.1597 & 0.2377 \\
DenseTNT~\cite{DBLP:journals/corr/abs-2108-09640} & 0.4195 & 1.1288 & 0.2240 \\
THOMAS~\cite{DBLP:journals/corr/abs-2109-01827}                & 0.4164 & 0.9679 & \textbf{0.1791} \\
\hline
\textbf{HDGT (Ours)}            & \textbf{0.3030} & \textbf{0.9580} & 0.1938
\end{tabular}
\caption{Multi-Agent Track}
\end{subtable}
\begin{subtable}{0.5\textwidth}
\centering
\begin{tabular}{llll}
 Method                     & minSADE$\downarrow$ & minSFDE$\downarrow$ & SMR$\downarrow$     \\ \hline
ReCoG2~\cite{mo2020recog} & 0.3295 & 0.8693 & 0.1498
 \\
THOMAS~\cite{DBLP:journals/corr/abs-2109-01827} & 0.3148 & \textbf{0.7162} & \textbf{0.1067} \\
DenseTNT~\cite{DBLP:journals/corr/abs-2108-09640} & 0.2786 & 0.8916 & 0.1502 \\
\hline
\textbf{HDGT (Ours)}            & \textbf{0.2255} & 0.7875 & 0.1322
\end{tabular}
\caption{Conditional Multi-Agent Track}
\end{subtable}
\label{tab:interaction-test}
\end{table*}

\begin{table}[!tb]
\centering

\caption{\textbf{Results on Waymo Motion Leaderboard, \textit{test} set.} }
\begin{tabular}{lllll}
                        & minADE$\downarrow$ & minFDE$\downarrow$ & MR$\downarrow$  & mAP$\uparrow$    \\ \hline
SimpleCNNOnRaster\cite{stepan2021mcnn}       & 0.7400 & 1.4936 & 0.2091        & 0.2136 \\
ReCoAt~\cite{hauang2021recoat}                  & 0.7703 & 1.6668 & 0.2437        & 0.2711 \\
DenseTNT~\cite{DBLP:journals/corr/abs-2108-09640}                & 1.0387 & 1.5514 & 0.1573       & 0.3281 \\
Kraken-NMS (Yandex SDG) & 0.6732 & 1.3947 & 0.1850      & \textbf{0.3646} \\
SceneTransformer~\cite{ngiam2021scene}        & 0.6117 & 1.2116 & 0.1564      & 0.2788 \\
$^{*}$MultiPath++~\cite{varadarajan2021multipath++}            & \textit{0.5557} & \textit{1.1577} & \textit{0.1340}     & \textit{0.4092} \\
\hline
\textbf{HDGT} (ours) & \textbf{0.5933}      &   \textbf{1.2055}           &         \textbf{0.1511}     &   0.2854    \\
\end{tabular}
\label{tab:waymo-test}
\end{table}

\begin{table*}[tb!]
\centering
\caption{\textbf{Results of Model Component Ablation Study.} }
 \small
      {
\begin{tabular}{ccccc|ccc}
\hline
Model variants& Relative & HD-Map & Semantic & Heterogeneous &  minADE $\downarrow$ & minFDE $\downarrow$& MR $\downarrow$ \\ \hline
 1&  &        &          &       &         0.2287 & 0.5338 & 0.0153 \\
2&   \Checkmark&        &          &        &        0.1889 & 0.4572 & 0.0093 \\
  3&  \Checkmark&        \Checkmark&          &        &       0.1526 & 0.3726 & 0.0044 \\
   4&  \Checkmark&        \Checkmark&          \Checkmark&        &       0.1328 & 0.3379 & 0.0038 \\
5 &   \Checkmark&        \Checkmark&          \Checkmark&        \Checkmark&      \textbf{0.1071} & \textbf{0.2945} & \textbf{0.0023} \\
\hline
\end{tabular}
}
\label{tab:ablation}
\end{table*}

\subsubsection{Implementation Details} 

The model is implemented by Pytorch and DGL. We use a hidden dimension of size 128. We use three Heterogeneous Driving Graph Transformer Layers on INTERACTION, similar to the number of GNN layers in \cite{mo2020recog,DBLP:journals/corr/abs-2109-06446}. On Waymo Open Motion, we have six lHeterogeneous Driving Graph Transformer Layers  (12.1 Million parameters) containing the same magnitude of parameters with \cite{ngiam2021scene} (15.3 Million parameters). Following the popular recipe for Transformer-based model, we use the AdamW optimizer with initial learning rate 5e-4, wight decay 1e-4, and batch size of 64. For both datasets, the number of training epochs is 30 with 1 epoch warmup and linearly decay to 0. The type-specific agent distance buffer hyper-parameter is: vehicle - 30 meters, pedestrian - 10 meters, cyclist - 20 meters, which is set empericallly. For those agents with missed time-step (marked with empty), we interpolate the trajectory if it is in the observed horizon and mask its regression loss if it is in the future horizon. Different from the original Transformer~\cite{vaswani2017attention}, we do not set dropout since it slows down the converge speed with no explicit performance gain. As for the data augmentation, agent drop is adopted~\cite{ngiam2021scene}, which randomly drop 10\% agents of a scene during training.  For ResNet-like 1D CNN, similar to~\cite{LiangECCV20}, we have three stages with each stage have two blocks. Different from~\cite{LiangECCV20}, we do not adopt an additional FPN as we observe no performance gain.  We use two layers of PointNet for polylines and polygons, similar to~\cite{Gao_2020_CVPR}. All experiments are conducted on NVIDIA Tesla V100.

\subsection{Compared Baselines}
\label{sec:compare_baselines}
We briefly introduce the methods we compare with on the public leaderboard of INTERACTION and Waymo Motion.  
\begin{itemize}
    \item \textbf{DenseTNT}~\cite{DBLP:journals/corr/abs-2108-09640} adopts VectorNet~\cite{Gao_2020_CVPR} as the encoder which models the scene as a homogeneous fully connected graph and proposes an optimization-based technique to select the best goal point from a predicted dense candidate set. Compared to direct decode trajectory, their approach could generate more diverse modes and thus have better MR.

  \item \textbf{MultiModalTransformer}~\cite{DBLP:journals/corr/abs-2109-06446} builds two fully connected graph (AgentAgent and AgentMap) and adopts the Transformer to encode the two graphs. The decoder is classification plus regression MLP (same with ours).

  \item \textbf{GOHOME}~\cite{DBLP:journals/corr/abs-2109-01827} encodes the scene with a LaneGCN~\cite{LiangECCV20}-like, which builds a series of four graphs (LaneToAgent, LaneToLane, LaneToActor, and finally ActorToActor) and extracts information with GCN~\cite{DBLP:conf/iclr/KipfW17}. Based on the graphs, they generate heatmap outputs representing the future position probability distribution of agents. Similar to DenseTNT, they could generate more diverse multi-modal predictions by an additional optimization step.

  \item \textbf{ReCoG2}~\cite{mo2020recog} builds a heterogeneous graph contains two types of nodes: vehicles and lanes, which is first encoded by RNN and CNN respectively. Then, they adopt GCN~\cite{DBLP:conf/iclr/KipfW17} for the graph and uses a RNN as decoder. Compared to HDGT, their map is encoded in the rasterized way and the GCN they used did not consider the heterogeneity of nodes.

  \item 
\textbf{THOMAS}~\cite{DBLP:journals/corr/abs-2109-01827}: extends GOHOME in a hierarchical way of heatmap generation for fine-grained prediction and an attention-based decoder for consistent multi-agent prediction.

  \item \textbf{SimpleCNNOnRaster}~\cite{stepan2021mcnn} and \textbf{ReCoAt}~\cite{hauang2021recoat} adopt CNN on rasterized images to encode map while~\cite{stepan2021mcnn} adopts rasterization for agents as well and ~\cite{hauang2021recoat} adopts RNN for agents. Their decoders are both classification plus regression MLP.

  \item 
  \textbf{SceneTransformer}~\cite{ngiam2021scene} builds a fully connected graph for all agents and map elements and they adopt Transformer on both spatial and temporal axis in an factorized way.

  \item \textbf{Multipath++}~\cite{varadarajan2021multipath++} is a concurrent work which extract agents' surrounding elements in their relative reference in encoder and proposes an ensemble method for decoding.

\end{itemize}

\subsection{Overall Performance Evaluation}\label{sec:sota}
Throughout 11/3/2021 (as of date for our final submission to INTERACTION public evaluation) to 4/21/2022 (as of the date for our initial submission of this paper), in terms of minADE/minFDE metric, we rank the  \textbf{best} result in all four tracks on the INTERACTION Leaderboard\footnote{http://challenge.interaction-dataset.com/prediction-challenge/intro}  and  \textbf{second}  on the Waymo Open Motion Leaderboard\footnote{https://waymo.com/open/challenges/2021/motion-prediction/}. Table~\ref{tab:interaction-test} and Table~\ref{tab:waymo-test} report the detailed results and comparison to previous state-of-the-arts.

\subsubsection{Results on INTERACTION}
Table~\ref{tab:interaction-test} shows that HDGT outperforms other candidates by a large margin in terms of minADE/minFDE.
Specifically, compared to MultiModalTransformer~\cite{DBLP:journals/corr/abs-2109-06446} and ReCoG2~\cite{mo2020recog} which has two seperate graphs and fully connected graph, HDGT outperforms them in all metrics, which demonstrates the importance of modelling symmetric relativity and heterogeneity as in HDGT.
As for the MR metric, we notice that DenseTNT and GOHOME has better ranks compared to their minADE/minFDE rank, which shows the effeciveness of their optimization-based decoding
THOMAS has the best MR in the multi-agent settings, which comes from their consistent multi-agent trajectory decoder.
Note that HDGT's output head is just the simple regression plus classification MLP scheme, which still has competitive MR results and thus demonstrates the strong representation vector from the encoder.

\subsubsection{Results on Waymo Open Motion} 
In Table~\ref{tab:waymo-test}, $^{*}$ denotes that the concurrent MultiPath++ involves ensemble with results shown in \textit{italic}; our HDGT ranks the first in the single model in terms of minADE/minFDE/MR as shown in \textbf{bold}.

We can observe that models with rasterized images for map encoding including SimpleCNNOnRaster~\cite{stepan2021mcnn} and ReCoAt~\cite{hauang2021recoat} have relative large performance gap compared to other methods, which verify the advantages of vector-based methods over raster-based methods. 
Again, DenseTNT~\cite{DBLP:journals/corr/abs-2108-09640} demonstrates strong performance on predicting agents' intentions, i.e., high MR and mAP while it has low scores on disntance-based metrics - minADE/minFDE, which is a widely known property of goal-based methods.
As for the SceneTransformer which adopts Transformer for scene encoding as well, HDGT outperforms it with lower cost. From the model complexity perspective, SceneTransformer has 15.3 million parameters while the submmited version of HDGT has 12.1 million parameters. From the computation efficiency perspective, SceneTransformer conducts attention mechanism on both spatial and temporal axes with two factorized fully connected graphs while HDGT only operates on one single heterogeneous graph sparesely connected according to semantic relations.
As for the concurrent work Multipath++, we observe its strong performance, which show the effectiveness of their ensemble technique.

In summary, HDGT, as a backbone for driving scene encoding, could outperform state-of-the-art backbones on L2-norm based metrics with simple output heads. For other types of metrics like MR/mAP, more advanced decoder as well as techniques like ensemble and test-time-augmentation could be explored.

\subsection{Ablation Study}\label{sec:ablation}

We conduct the ablation study to validate the effectiveness of each design in HDGT and the results are shown in Table \ref{tab:ablation}.
The experiments are conducted on the validation set of INTERACTION dataset and we keep all the other settings the same. Note that all models have enough capacities to overfit the training set and we report their best performance on the validation set. 
In Table~\ref{tab:ablation}, \emph{Relative} means using local coordinate system for each node and its in-edges, otherwise we set the autonomous vehicle as the global reference. \emph{HD-Map} means using HD-Map otherwise only agents' information. \emph{Semantic} means building graph according to distance threshold and lane connection information otherwise building a fully-connected graph. \emph{Heterogeneous} means using different parameters for
 different node and edge types otherwise sharing the parameters.
From the results, we can conclude that:
\begin{itemize}[topsep=0ex,itemsep=0ex]
    \item \textit{Relative representation in the allocentric view is better.} In model 1, we set the autonomous vehicle as the global reference while in model 2 we adopt the proposed node-centric local coordinate system. The improvement of the model 2 shows the necessity of learning relative spatial relations instead of absolute coordinates, which aligns well with the conclusions in ~\cite{jia2021multiagent, mo2021heterogeneous, varadarajan2021multipath++}.
    \item  \textit{Road semantics from HD-map are important.} In model 3, we add HD-Map and construct a homogeneous fully-connected graph similar to VectorNet~\cite{Gao_2020_CVPR}. It has better performance compared to model 2, which proves the importance of HD-map information for the trajectory prediction. For example, drivers tend to drive following the lane centerline. When it comes to parking, they would stop nearby the curb.
    \item \textit{Semantic relations are helpful.} In model 4, instead of a fully-connected graph, we build the semantic graph as proposed in HDGT but the GNN parameters for different types of nodes are still shared. As a result, it brings improvement compared to model 3 while worse than model 5.
    \item \textit{Heterogeneous graph network parameters are vital.} Model 5 is the full version of HDGT and we can find that specific parameters for different types of nodes and edges could boost the performance significantly compared to model 4.
\end{itemize}

\begin{table*}[h!]
\centering
\caption{\textbf{Results of Different Relative Encoding Strategies.} Averaged over 3 runs.}
      {
\begin{tabular}{c|ccc}
\hline
Relative Encoding  & minADE $\downarrow$ & minFDE $\downarrow$& MR $\downarrow$ \\ \hline
 Fixed Reference &  0.1549$\pm$0.0032 & 0.4079$\pm$0.0095 & 0.0051$\pm$0.0000 \\
Relative Edge &  0.1158$\pm$0.0021 & 0.3296$\pm$0.0060 & 0.0025$\pm$0.0000 \\
Pose Change (Ours) &  \textbf{0.1081}$\pm$0.0030 & \textbf{0.2945}$\pm$0.0061 & \textbf{0.0023}$\pm$0.0000 \\
\hline
\end{tabular}
}
\vspace{-2mm}
\label{tab:relative-encoding}
\end{table*}

\begin{table*}[tb!]
\centering
\caption{\textbf{Performance by Different Numbers of Heterogeneous Driving Graph Transformer Layers.}}
\begin{tabular}{c|cccc|cccc}
Dataset &  \multicolumn{4}{c}{INTERACTION}  & \multicolumn{4}{c}{Waymo}         \\ \hline
  Layer \# & minADE $\downarrow$ & minFDE $\downarrow$ & MR $\downarrow$ & Inference Time (second)   & minADE $\downarrow$ & minFDE $\downarrow$ & MR $\downarrow$ & Inference Time (second)\\ \hline  
1             & 0.1430 & 0.3997 & 0.0043  & 95 & 0.6462& 1.3099 & 0.1865 & 901  \\
2             & 0.1149 & 0.3243 & 0.0025 & 100  & 0.6013 & 1.2004  & 0.1648 & 977 \\
3             & \textbf{0.1071}&\textbf{0.2945} & 0.0023 & 107  & 0.5835 & 1.1736 & 0.1601 & 1038 \\
6             & 0.1082 & 0.3035 & \textbf{0.0022} & 122 &  \textbf{0.5673}&\textbf{1.1507} & \textbf{0.1505} & 1146 \\
\hline
\end{tabular}
\vspace{-5mm}
\label{tab:Scalability}
\end{table*}

\begin{table}[tb!]
\centering
\caption{\textbf{Results of Heterogeneous Graph Deisgn.}}
 \small
      {
\begin{tabular}{c|ccc}
\hline
Model   & minADE $\downarrow$ & minFDE $\downarrow$& MR $\downarrow$ \\ \hline
HDGT (Ours) &  \textbf{0.1071} & \textbf{0.2945} & \textbf{0.0023} \\
Merge Lane Connectivity &  0.1622 & 0.4003 & 0.0078 \\
Homogeneous Map Node &  0.1272 & 0.3171 & 0.0031 \\
GCN Function &  0.1882 & 0.4721 & 0.0114 \\
\hline
\end{tabular}
\vspace{-5mm}
}
\label{tab:graph-design}
\end{table}

\subsection{Comparison of Relative Encoding Strategy\label{sec:comparision-relative}}
In Sec.~\ref{sec:relative}, we discuss the importance to encode the agent in the relative reference. Here, we compare three strategies to model the relativity among agents:
\begin{enumerate}
    \item \textbf{Fixed Reference}: it select one agent (usually the ego agent) as the global reference, i.e, adopting its coordinate and heading at the last observed time-step as the origin and positive x-axis of the global coordinate system. It is widely used in VectorNet~\cite{Gao_2020_CVPR} series of work, LaneGCN~\cite{LiangECCV20} series of work, and SceneTransformer~\cite{ngiam2021scene}. This strategy could achieve decent performance for the ego agent. However, since other agents are not in their local reference system, people have to forward model \textit{number of agents} times with each agent as the gloabal reference to achieve the best performance for multi-agent prediction~\cite{ettinger2021large}.
    \item \textbf{Relative Edge}: One modification of Fixed reference is to have each agent in their local coordinate system with in-edges also transformed into its target nodes' coordinate system. It is adopted in methods including concurrent work Multipath++~\cite{varadarajan2021multipath++} and HEAT-I-R~\cite{mo2021heterogeneous}. One potential downside of this strategy is that for those in-edges with the same source node, their features are encoded separately which ignores the fact that they are indeed one object.
    \item \textbf{Pose Change}: to this end, we encode the relative feature by combing the source nodes' feature with $\Delta\text{Pose}$. The concurrent work HiVT also designs a Global Interaction Module which encodes the pair-wise coordinate transformation. 
\end{enumerate}

We conduct experiments with the above three strategies on the validation set of INTERACTION dataset, while keeping other designs of HDGT unchanged. The results are in Table~\ref{tab:relative-encoding} and we can find that fixing reference has explicit performance compared to encoding each agent in their own coordinate system. When it comes to the Relative Edge and Pose Change, we can observe that Pose Change demonstrates better performance, verifying the effectiveness of the proposed Relative Relation Encoding technique.

\subsection{Analysis of Heterogeneous Graph Design\label{sec:het-graph-design}}
In this section, we examine the following design choice:
\begin{itemize}
    \item \textbf{Merge Lane Connectivity}: we merge the four types of lane connectivity edges into one to examine the necessity of the modeling the lane topology with separate parameters.
    \item \textbf{Homogeneous Map Node}: we merge the node type of all lanes and traffic signs while keep all the edges unchanged to examine the necessity to process lanes and traffic signs with different parameters.
    \item \textbf{GCN Function}: we adopt the GCN~\cite{DBLP:conf/iclr/KipfW17} like aggregation and update function as in LaneGCN~\cite{LiangECCV20} to compare with Transformer in HDGT. The in-edges' features of each node are fed into an MLP and then an summation operation is conducted to aggregate information. For update function of nodes, we concatenate the aggregated information with the old node features and feed them into an MLP. For different node and edge types, we still adopt different parameters.
\end{itemize}
We compare the three aforementioned design choice with the proposed one on the validation set of INTERACTION dataset. The results are in Table~\ref{tab:graph-design} and we can conclude that: (i) Change the Transformer to GCN-like function leads to the largest performance drop, which demonstrates the advantages of adopting Transformer structure. (ii) Ignoring the lane topology leads to the second largest performance drop. The result is  in accordance with expectation since vehicles tends to move along the lane and lane topology could provide useful information to reduce the uncertainly of the prediction, which aligns with the conclusions in existing works~\cite{LiangECCV20,deo2021multimodal}. (iii) Merge the lane and traffic signs leads to the least performance drop. It might come from the fact during the initialization of the feature, the sub-type information has been encoded by learnable embeddings, which could be used to distinguish lanes and traffic signs. However, compared to the original design, it still leads to performance drop which might be because of the different characteristics of the two types of objects.

\subsection{Exploration on Model Capacity}\label{sec:scale}
We examine the scalability of HDGT. We conducted experiments on the validation set of INTERACTION and Waymo Open Motion by models with different number of  Heterogeneous Driving Graph Transformer Layers. Additionally, we report the inference time on the each datasets' validation set for 1 epoch to estimate the influence of different number of layers. 
For Waymo, the minADE/minFDE are calculated in their official way.
From Table ~\ref{tab:Scalability}, we can observe that for INTERACTION (40K tracks), the performance saturates with around 3 Heterogeneous Driving Graph Transformer Layers. When it comes to the larger Waymo Open Motion (7.64M tracks), the performance could be still improved even with 6 layers. 
This shows that HDGT enjoys the strong scalability of Transformer structure, which brings huge success in the vision~\cite{zhai2021scaling} and language~\cite{brown2020language} fields. The simple and yet unified structure of HDGT makes it easy to extend its capacity when there is more data, which fits perfectly for the autonomous driving industry.

\begin{figure*}[!tb]
		\centering
		\begin{subfigure}[t]{0.45\textwidth}
			\centering
		\captionsetup{width=8.0cm}
		\includegraphics[width=8.0cm]{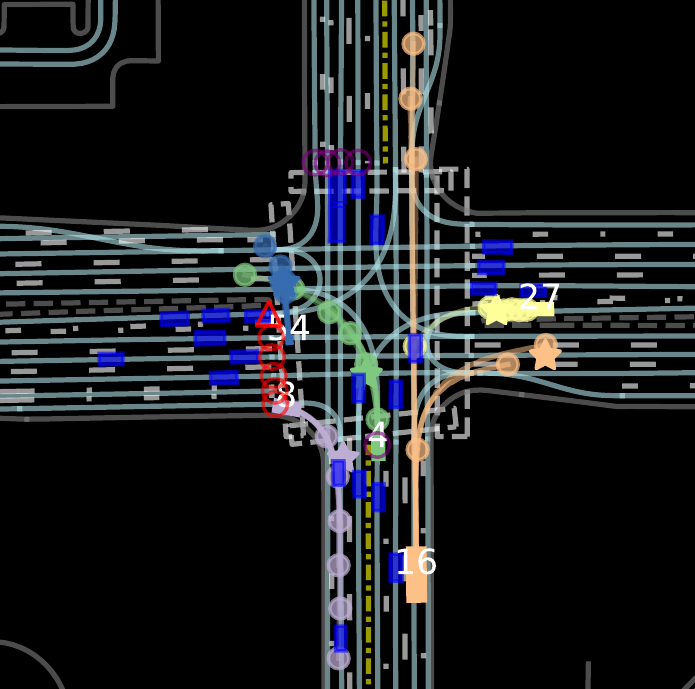}
		\captionsetup{width=8.0cm}
			\caption{\textbf{P54} was passing the road through the crosswalk while \textbf{V4} desired to turn left. The model outputs their different potential futures. For \textbf{V16}, as its target is unclear by observation, the model outputs diverse futures: go straight, yield, or right turn.}
		\end{subfigure}%
		\begin{subfigure}[t]{0.45\textwidth}
			\centering
	   	\includegraphics[width=8.0cm]{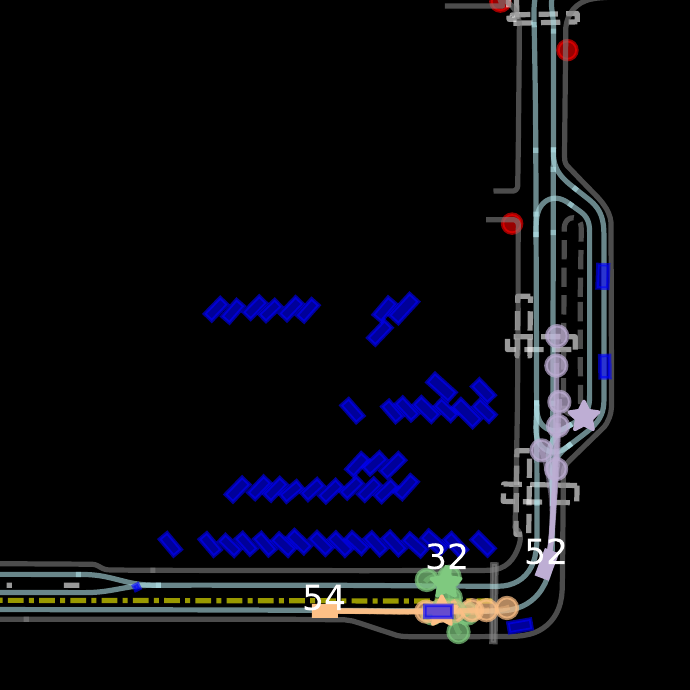}
		\captionsetup{width=8.0cm}
			\caption{Since \textbf{P32} could either walk along curb or pass the road, the model outputs the proper reactions of \textbf{V54} under both conditions. For \textbf{V52}, its possibility for going at different lanes are all considered.}
		\end{subfigure}%
		\\
		\begin{subfigure}[t]{0.45\textwidth}
			\centering
			\includegraphics[width=8.0cm]{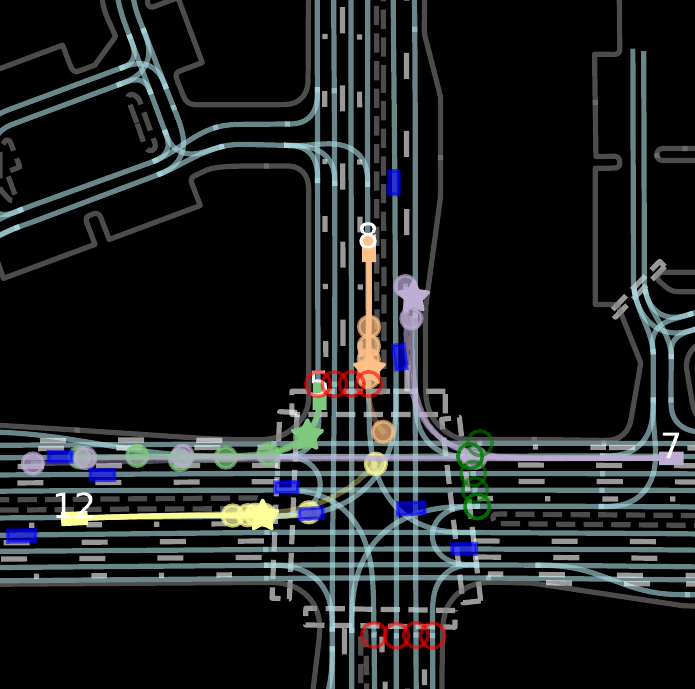}
			\captionsetup{width=8.0cm}
			\caption{\textbf{P7} could either go straight through the green light or turn right. Thus, the prediction of \textbf{P5} is influenced. For \textbf{V8} and \textbf{V12}, they were far away from the traffic light and thus their predictions include both passing and yielding.}
		\end{subfigure}%
		\begin{subfigure}[t]{0.45\textwidth}
			\centering
		\captionsetup{width=8.0cm}
		\includegraphics[width=8.0cm]{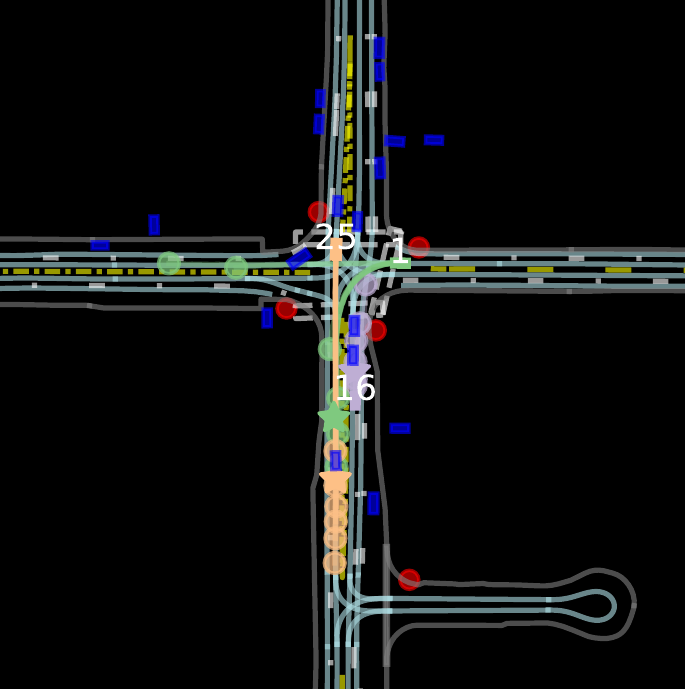}
			\caption{\textbf{V1}'s future had high uncertainty (going straight or left turn) and it was also influenced by either \textbf{V25} or \textbf{V16} under different situations. Also, the existing of stop sign (red double ring) results in conservative predictions for \textbf{V16}.}
		\end{subfigure}%
		\caption{\textbf{Visualization of Prediction Results on Waymo Opem Motion Dataset.} $\star$ represents the ground-truth final position and $\circ$ denotes predicted final positions of modes (lower confidence, more transparent). Abbreviation: V-vehicle, P-pedestrian, and C-cyclist.}
\label{fig:vis}
\end{figure*}

\subsection{Visualization} \label{sec:visualize}
We visualize the prediction results of HDGT in some scenarios on the Waymo Open Motion dataset in Fig.~\ref{fig:vis}. One can find that to predict agents' future trajectories accurately, information about diverse elements and their relations are necessary such as agents, lanes, stop-signs, traffic lights and \textit{etc}.  Thus, the explicit modeling of the heterogeneous nature of the driving scene in HDGT is significant and beneficial for the trajectory prediction.

\section{Conclusion}
We have proposed a principled approach to extract better representation for trajectory prediction. It models the driving scene as a heterogeneous graph and adapt transformer as the graph aggregation and update function. 

The spatial features are normalized into the local coordinate system when being aggregated. The proposed approach has achieved state-of-the-art results on two recent large-scale and competitive benchmarks. Thorough ablation studies validate the effectiveness of each module in our method.

\section*{Acknowledgement}
This work was partly supported by NSFC (62222607,62206172), Shanghai Municipal Science and Technology Major Project (2021SHZDZX0102). The authors are thankful to the anonymous reviewers for their valuable comments.

\bibliographystyle{IEEEtran}
\bibliography{ref}

\begin{IEEEbiography}
[{\includegraphics[width=1in,height=1.25in,clip,keepaspectratio]{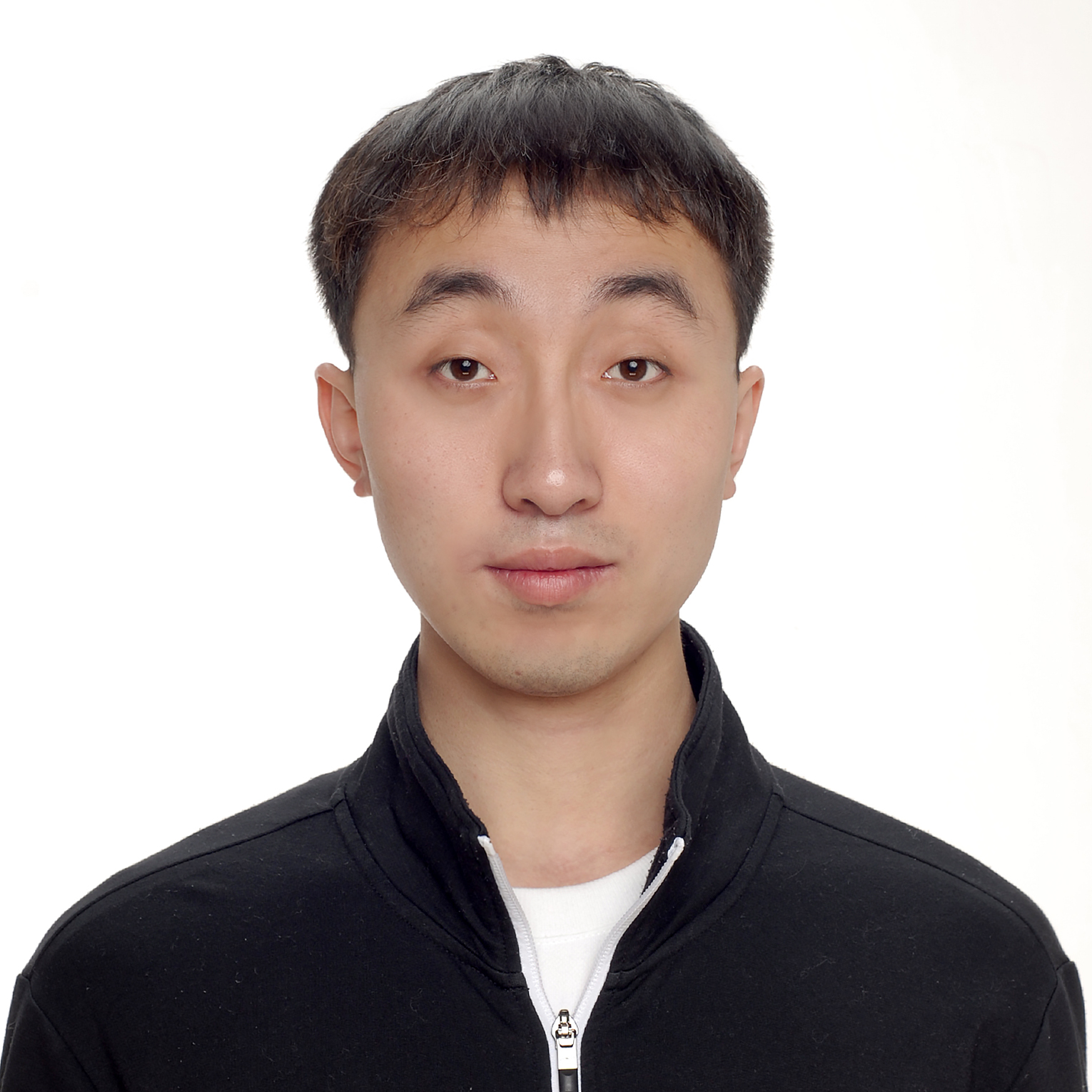}}]
	{Xiaosong Jia} is currently a PhD student at Department of Computer Science and Engineering, Shanghai Jiao Tong University (SJTU), Shanghai. Before that, he earned B.E. in IEEE Honor class at SJTU. His research interests include autonomous driving and machine learning, with (co-) first-authored papers published in CoRL, NeurIPS, CVPR, RAL, TKDE, etc.
\end{IEEEbiography}

\begin{IEEEbiography}
[{\includegraphics[width=1in,height=1.25in,clip,keepaspectratio]{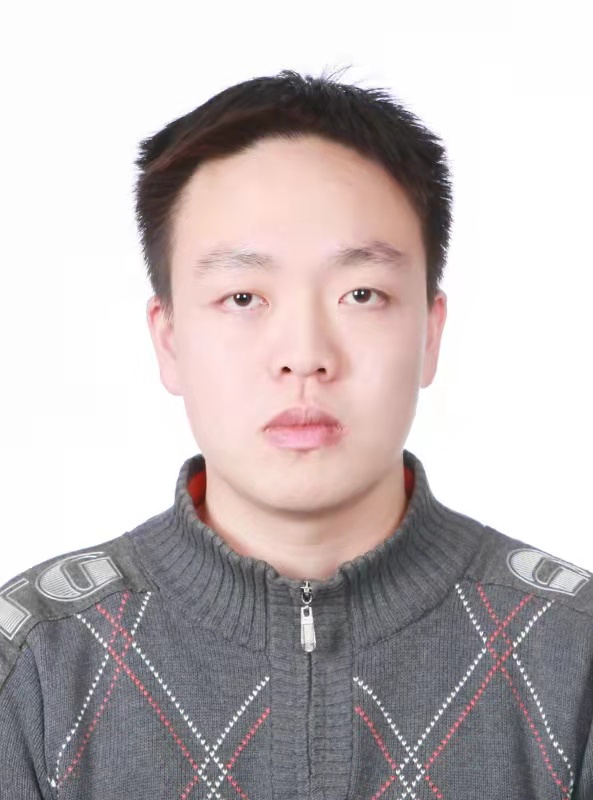}}] {Penghao Wu} 

was once an undergraduate student with the University of Michigan-Shanghai Jiao Tong University Joint Institute (UM-SJTU JI), Shanghai Jiao Tong University (SJTU), Shanghai where he obtained his B.E. in electrical and computer engineering. He is now a master student with department of computer science, UCSD, USA. His research interests include autonomous driving and computer vision. He has published first/co-authored paper in ECCV, NeurIPS, ICLR on autonomous driving.
\end{IEEEbiography}

\begin{IEEEbiography}
	[{\includegraphics[width=1in,height=1.25in,clip,keepaspectratio]{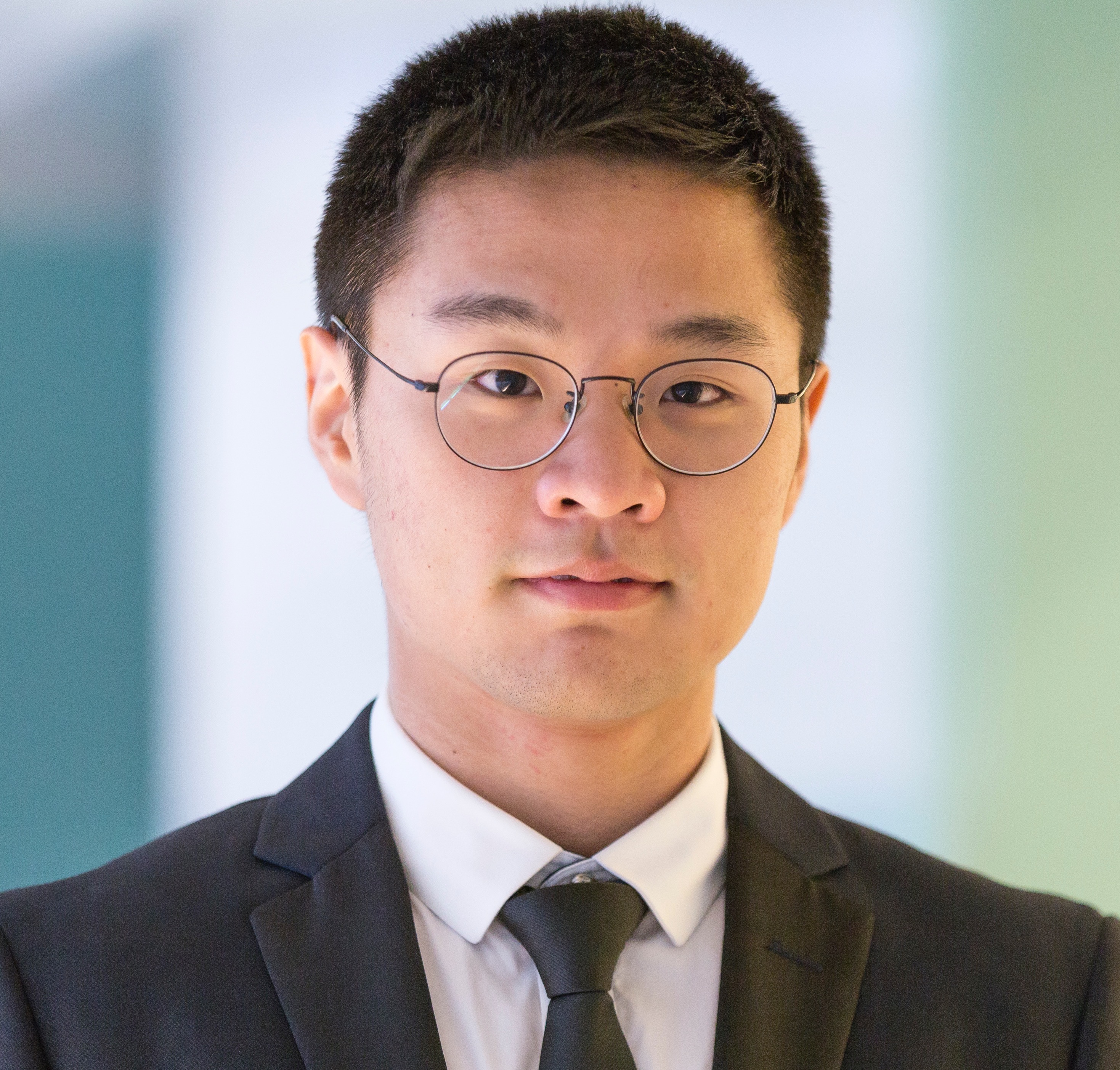}}]
	{Li Chen} received the B.E. in mechanical engineering from Shanghai Jiao Tong University in 2019, and the M.S. in Robotics from University of Michigan, Ann Arbor, USA in 2020. He was once a research engineer with SenseTime after graduation, and is currently a researcher with Shanghai AI Laboratory, Shanghai, China. His research interests include autonomous driving, computer vision and machine learning. He has published first/co-authored paper in ECCV, CoRL, NeurIPS on autonomous driving.
\end{IEEEbiography}

\begin{IEEEbiography}[{\includegraphics[width=1in,height=1.25in,clip,keepaspectratio]{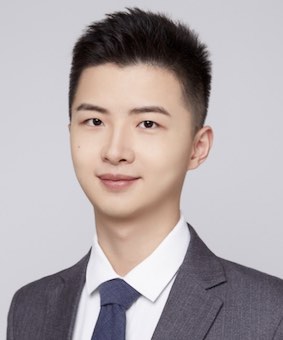}}]{Yu Liu} is a Principal Investigator in Shanghai AI Lab and also the General Manager of ADG Business Unit and a Senior Director of research in SenseTime Group, Hong Kong. Before that, he obtained the PhD from Multimedia Lab (MMLab), the Chinese University of Hong Kong, and B.E. from Beihang University, Beijing, China in 2016. His research interests include Artificial General Intelligence, Deep RL and neural network understanding with big data.\end{IEEEbiography}

\begin{IEEEbiography}[{\includegraphics[width=1in,height=1.25in,clip,keepaspectratio]{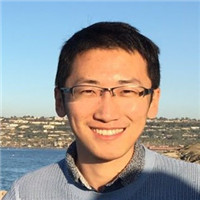}}]{Hongyang Li} received PhD in computer science from Chinese University of Hong Kong, in 2020, and the B.E. degree in Computer Science from Dalian university of technology in 2014. He is currently a Research Scientist and leading the OpenDriveLab team  at Shanghai AI Lab. His expertise focuses on perception and cognition, end-to-end autonomous driving and foundation model. He is also affiliated with Department of Computer Science and Engineering, Shanghai Jiao Tong University as a post-doc researcher.
\end{IEEEbiography}

\begin{IEEEbiography}
[{\includegraphics[width=1in,height=1.25in,clip,keepaspectratio]{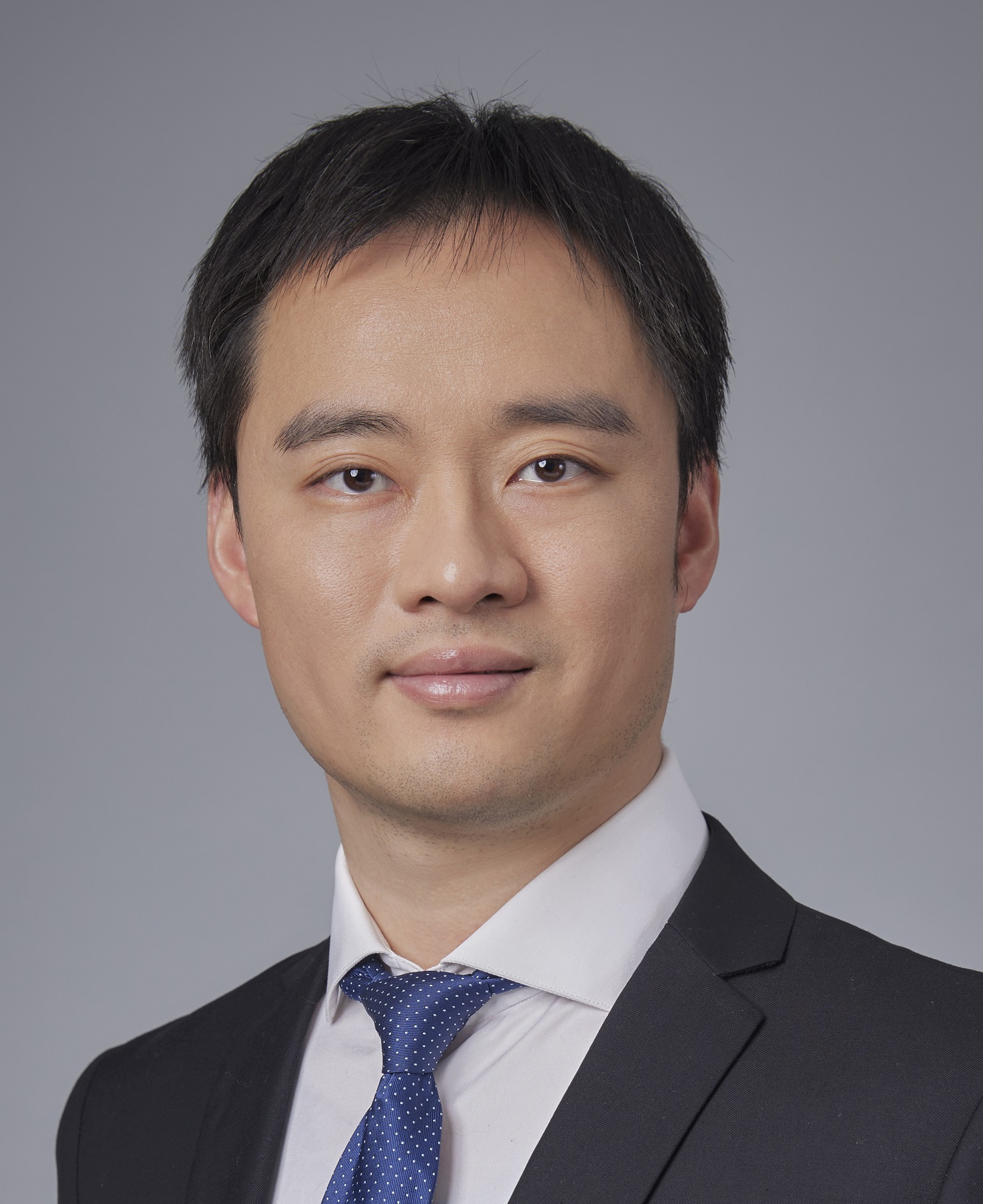}}] 
{Junchi Yan} (S'10-M'11-SM'21) is currently an Associate Professor with Department of Computer Science and Engineering, Shanghai Jiao Tong University, Shanghai, China. Before that, he was a Senior Research Staff Member with IBM Research where he started his career since April 2011. His research interests include machine learning and computer vision. He regularly serves as Senior PC/Area Chair for NeurIPS, ICML, CVPR, AAAI, IJCAI, ACM-MM and Associate Editor for the Pattern Recognition Journal.
\end{IEEEbiography}

\end{document}